\documentclass[runningheads]{llncs}
\usepackage[T1]{fontenc}
\usepackage{graphicx}
\usepackage{hyperref}
\usepackage{color}

\usepackage{subcaption}
\usepackage{amsmath}
\usepackage[table]{xcolor}
\usepackage{cleveref}
\usepackage{tcolorbox}
\tcbuselibrary{listings,breakable}

\usepackage[utf8]{inputenc}
\usepackage[T1]{fontenc}
\usepackage{lmodern}
\usepackage{caption}
\usepackage{longtable}
\usepackage{booktabs}
\usepackage{siunitx}
\usepackage{tabularx}    
\usepackage{multirow}    

\usepackage{listings}
\usepackage{enumitem}

\usepackage{tikz}
\usetikzlibrary {shapes.geometric}
\usetikzlibrary {fit} 

\usepackage{pgfplots}
\usepgfplotslibrary{statistics}

\pgfplotsset{
    compat=1.3,
    chatclimate pure/.style={
        boxplot,
        ultra thick,
        draw=darkgray,
        solid,
        fill=darkgray!15,
        mark=*,
        every mark/.append style={
            fill=darkgray!15,
        },
    },
    chatclimate ecoscapes/.style={
        boxplot,
        ultra thick,
        draw=green,
        solid,
        fill=green!15,
        mark=*,
        every mark/.append style={
            fill=green!15,
        },
    },
}

\begin{document}
\title{EcoScapes: LLM-Powered Advice for Crafting Sustainable Cities}
%
%
\author{Martin Röhn\inst{1,2}\orcidID{0009-0008-9475-7851} \and
Nora Gourmelon\inst{2}\orcidID{0000-0003-3760-0184} \and
Vincent Christlein\inst{2}\orcidID{0000-0003-0455-3799}}
%
%
\institute{Technische Universität Nürnberg, 90461 Nürnberg, Germany \and
Friedrich-Alexander-Universität Erlangen-Nürnberg, 91058 Erlangen, Germany
\email{martin.roehn@utn.de}
\email{\{nora.gourmelon,vincent.christlein\}@fau.de}}
\maketitle              
\begin{abstract}
Climate adaptation is vital for the sustainability and sometimes the mere survival of our urban areas.
However, small cities often struggle with limited personnel resources and integrating vast amounts of data from multiple sources for a comprehensive analysis.
To overcome these challenges, this paper proposes a multi-layered system combining specialized LLMs, satellite imagery analysis and a knowledge base to aid in developing effective climate adaptation strategies.
The corresponding code can be found at \url{https://github.com/Photon-GitHub/EcoScapes}.

\end{abstract}
\section{Introduction}
\subsection{Importance of Climate Adaptation}
Climate change poses the largest environmental and societal challenge of our time, with its impacts increasingly manifested in frequent and severe weather events, including heatwaves, droughts, and floods.
These phenomena threaten the viability and sustainability of urban areas worldwide~\cite{ipcc2023FullReport}.
Effective climate adaptation is therefore critical, aimed at preserving urban functionality and mitigating disaster impacts.
However, the Intergovernmental Panel on Climate Change (IPCC) identifies significant constraints for smaller cities, including limited financial resources and administrative capacities, categorizing these as "soft adaptation limits" already reached in various sectors and regions~\cite{ipcc2023FullReport}.

\subsection{AI in Climate Adaptation Planning}
Recent advances in Artificial Intelligence (AI), particularly large language models (LLMs), offer new opportunities to simplify climate adaptation planning.
Transformer-based models such as GPT have revolutionized natural language processing through advanced tokenization methods, positional encodings, and attention mechanisms, allowing these models to generate contextually rich human-like text~\cite{vaswani2017attention,radford2018PreTrainedKnowledge}. 
Despite significant advancements exemplified by ChatGPT and its GPT-4 model~\cite{bubeck2023sparks,openai2024gpt4technicalreport}, LLMs remain susceptible to inaccuracies known as hallucinations, presenting risks in critical application domains such as climate science.

To mitigate inaccuracies in LLM outputs and enhance the reliability of climate adaptation strategies, systems like ChatClimate~\cite{vaghefi2023chatclimate} (CC) have emerged, integrating authoritative sources such as the IPCC's Sixth Assessment Report (IPCC AR6)~\cite{ipcc2023FullReport}.
CC significantly improves LLM-generated outputs by directly attributing responses to validated scientific sources, reducing instances of misinformation and significantly improving accuracy~\cite{vaghefi2023chatclimate}.
However, existing systems remain limited in their ability to deliver geographically specific recommendations due to their reliance on generalized knowledge bases.

\subsection{EcoScapes}
This paper introduces EcoScapes, a proof-of-concept decision-support framework aiming to extend existing AI systems like CC by incorporating multimodal LLMs to analyze local satellite imagery.
Utilizing data from the Sentinel-2 satellite program, EcoScapes attempts to analyze local urban characteristics such as land cover, water bodies, and ground moisture levels, to tailor advice for environmental risks, including heat islands and flood susceptibility.
This multimodal approach seeks to address critical gaps in local specificity and accuracy, although the presented results remain preliminary and exploratory.

Ultimately, this research aims to explore how AI-based tools could assist smaller communities by providing preliminary insights in form of a report to address resource limitations, reduce the complexity of initial planning stages, and foster more accessible engagement in proactive climate adaptation strategies.
\section{Satellite Images}

\subsection{Data Acquisition}
\begin{figure}[htb]
    \centering
    \begin{subfigure}[b]{0.3\linewidth}
        \centering
        \includegraphics[width=\linewidth]{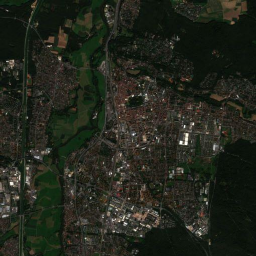}
        \caption{RGB}
    \label{fig:methodology:satellite_image_comparison:rgb}
    \end{subfigure}
    \hspace{0.02\linewidth}
    \begin{subfigure}[b]{0.3\linewidth}
        \centering
        \includegraphics[width=\linewidth]{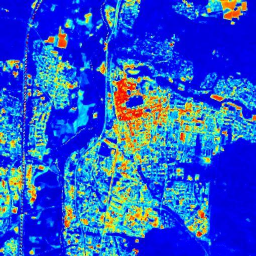}
        \caption{Moisture}
    \label{fig:methodology:satellite_image_comparison:moisture}
    \end{subfigure}
    \hspace{0.02\linewidth}
    \begin{subfigure}[b]{0.3\linewidth}
        \centering
        \includegraphics[width=\linewidth]{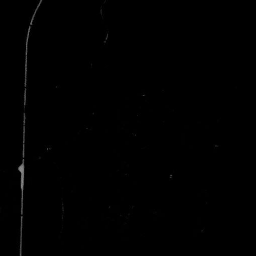}
        \caption{Water}
    \label{fig:methodology:satellite_image_comparison:water}
    \end{subfigure}
    \caption{Comparison of RGB and two satellite band indices. For Moisture, blue areas indicate high moisture and red areas low moisture levels. Water is a greyscale image. The channel was detected successfully, but some rivers were not.}
    \label{fig:methodology:satellite_image_comparison}
\end{figure}
The primary satellite imagery source used in this research is the Sentinel-2 mission, part of the European Space Agency's Copernicus Program.
Sentinel-2 satellites provide optical imagery in multiple spectral bands, effectively supporting environmental monitoring tasks such as water detection and soil moisture analysis, as shown in Figure \ref{fig:methodology:satellite_image_comparison}.

To acquire relevant imagery, we employed the SentinelHub service, which provides streamlined access to Sentinel-2 data via the EcoScapes framework’s Satellite Loader module.
This module retrieves images based on the geographical coordinates of targeted urban areas.

Automated access via SentinelHub, as implemented in EcoScapes, requires a paid subscription, but manual retrieval through the Copernicus Data Browser~\cite{copernicus_data_browser} is freely available. 
EcoScapes will skip accessing SentinelHub if manually added images are present.
Detailed manual retrieval instructions are provided in Appendix \ref{appendix:copernicus_data_browser}, ensuring accessibility regardless of budget constraints.

The Sentinel-2 imagery offers 13 distinct bands spanning visible to short-wave infrared spectra~\cite{copernicus_data_browser}. 
This study utilized the bands listed in Table~\ref{table:sentinel_bands}.
\begin{table}[htbp]
    \caption{Sentinel-2 bands and their applications}
    \label{table:sentinel_bands}
    \centering
    \begin{tabularx}{\linewidth}{l l S[table-format=4.1] X}
        \toprule
        \textbf{Band} & \textbf{Name} & \textbf{Wavelength (nm)} & \textbf{Use} \\
        \midrule
        B02 & Blue & 492.1 & RGB (Fig.~\ref{fig:methodology:satellite_image_comparison:rgb}) \\
        B03 & Green & 559.0 & RGB and Water (Figs.~\ref{fig:methodology:satellite_image_comparison:rgb},~\ref{fig:methodology:satellite_image_comparison:water}) \\
        B04 & Red & 665.0 & RGB (Fig.~\ref{fig:methodology:satellite_image_comparison:rgb}) \\
        B08 & Near-Infrared & 833.0 & Water (Fig.~\ref{fig:methodology:satellite_image_comparison:water}) \\
        B8A & Narrow Near-Infrared & 864.0 & Moisture (Fig.~\ref{fig:methodology:satellite_image_comparison:moisture}) \\
        B11 & Short-Wave Infrared & 1610.0 & Moisture (Fig.~\ref{fig:methodology:satellite_image_comparison:moisture}) \\
        \bottomrule
    \end{tabularx}
\end{table}

To calculate the moisture (Figure \ref{fig:methodology:satellite_image_comparison:moisture}) and water (Figure \ref{fig:methodology:satellite_image_comparison:water}) images, indices were used. 
These indices are calculated using a normalized difference formula:
\begin{equation}
\text{Index} = \frac{a - b}{a + b}
\end{equation}
where a and b are specific spectral bands chosen to enhance the visibility of terrain features.
For moisture, $a=B8A,b=B11$ and for water $a=B03,b=B08$.

\subsection{Limitations}
Despite the versatility of Sentinel-2 imagery, several notable limitations exist. Cloud cover hinders surface analysis, prompting the Satellite Loader to employ a strict filter that selects images with less than 1\,\% cloud cover from the preceding year. 
While effective, this strategy introduces potential biases towards drier conditions, potentially underrepresenting seasonal water bodies or overstating dryness and heat risks.

\begin{figure}[tb]
    \centering
    \begin{subfigure}[b]{0.45\linewidth}
        \centering
        \includegraphics[height=6cm]{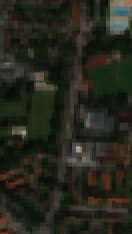}
        \caption{Zoomed-in image of a major arterial road}
    \end{subfigure}
    \hspace{0.05\linewidth}
    \begin{subfigure}[b]{0.45\linewidth}
        \centering
        \includegraphics[height=6cm]{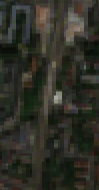}
        \caption{Zoomed-in image of a multi-track railway}
    \end{subfigure}
    \caption{Comparison of zoomed-in satellite images showing a street and a railway. Due to the limited resolution, it is difficult to distinguish between these two features.}
    \label{fig:methodology_limitations_resolution}
\end{figure}

Moreover, Sentinel-2's spatial resolution, although high relative to alternatives such as Landsat~\cite{usgs_landsat} and MODIS~\cite{nasa_modis}, remains insufficient for capturing detailed urban structures.
Consequently, even major infrastructure elements can be challenging to differentiate as illustrated in Figure~\ref{fig:methodology_limitations_resolution}.

Overall, despite these constraints, Sentinel-2 imagery remains an effective foundation for environmental analysis within the EcoScapes framework, balancing accessibility and analytical accuracy.

\section{Architecture}
\label{sec:architecture}

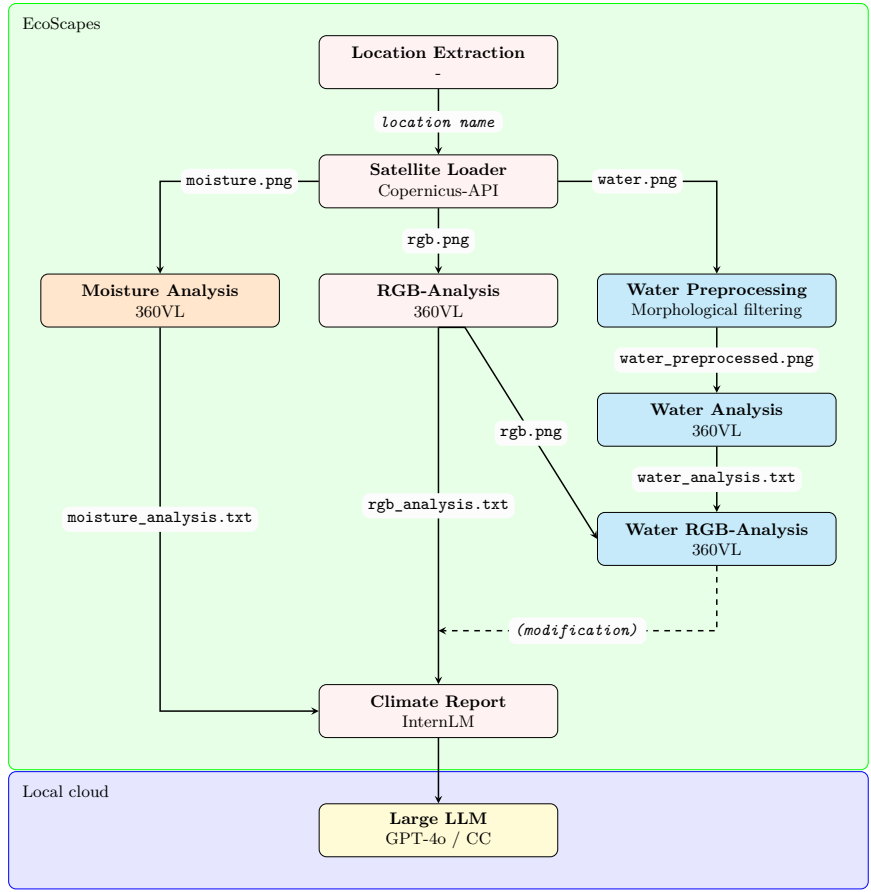
\begin{figure}[htb]
    \centering
    \resizebox{0.95\linewidth}{!}{
    \begin{tikzpicture}[node distance=2.25cm, every node/.style={fill=white, font=\small}]
        \tikzstyle{rgb} = [rectangle, rounded corners, minimum width=4.5cm, minimum height=1cm, draw=black, fill=pink!20, text centered, align=center, font=\footnotesize]
        
        \tikzstyle{water} = [rgb, fill=cyan!20]

        \tikzstyle{moisture} = [rgb, fill=orange!20]

        \tikzstyle{llm} = [rgb, fill=yellow!20]

        \tikzstyle{invisible} = [rgb, draw=none, fill=none]

        \tikzstyle{arrow} = [thick,->,>=stealth]
        \tikzstyle{soft_arrow} = [arrow, dashed]

        \tikzstyle{arrow_description} = [fill=lightgray!5, rounded corners, text centered, align=center, font=\footnotesize\ttfamily]
        \tikzstyle{background_box} = [rounded corners, minimum width=\linewidth, inner sep=0.6cm]

        \pgfdeclarelayer{background}
        \pgfsetlayers{background,main}

        \node (location) [rgb] {\textbf{Location Extraction} \\ -};
        \node (satellite) [rgb, below of=location] {\textbf{Satellite Loader} \\ Copernicus-API};

        \node (rgbanalysis) [rgb, below of=satellite] {\textbf{RGB-Analysis} \\ 360VL};
        \node (waterpreprocessing) [water, right of=rgbanalysis, xshift=3cm] {\textbf{Water Preprocessing} \\ Morphological filtering};
        \node (wateranalysis) [water, below of=waterpreprocessing] {\textbf{Water Analysis} \\ 360VL};
        \node (waterrgb) [water, below of=wateranalysis] {\textbf{Water RGB-Analysis} \\ 360VL};

        \node (moistureanalysis) [moisture, left of=rgbanalysis, xshift=-3cm] {\textbf{Moisture Analysis} \\ 360VL};
        \node (report) [rgb, below of=rgbanalysis, yshift=-5.5cm] {\textbf{Climate Report} \\ InternLM};

        \node (llm) [llm, below of=report] {\textbf{Large LLM} \\ GPT-4o / CC};

        \node (llmleftinvis) [invisible, left of=llm, xshift=-3cm] {};
        \node (llmrightinvis) [invisible, right of=llm, xshift=3cm] {};
        
        \begin{pgfonlayer}{background}
            \node (ecoscapesbox) [background_box, draw=green, fill=green!10, fit={(location.north) (report.south) (moistureanalysis.west) (wateranalysis.east)}, label={[anchor=north west, yshift=-0.15cm, xshift=0.15cm]north west:EcoScapes}] {};
            
            \node[background_box, draw=blue, fill=blue!10, fit={(llm) (llmleftinvis.west) (llmrightinvis.east)}, label={[anchor=north west, yshift=-0.15cm, xshift=0.15cm]north west:Local cloud}] {};
        \end{pgfonlayer}

        \draw [arrow] (location) -- node [arrow_description] {\textit{location name}} (satellite);
        \draw [arrow] (satellite) -- node [arrow_description] {rgb.png} (rgbanalysis);
        \draw [arrow] (satellite) -| node [arrow_description, near start] {water.png} (waterpreprocessing);
        \draw [arrow] (satellite) -| node [arrow_description, near start] {moisture.png} (moistureanalysis);

        \draw [arrow] (rgbanalysis.south) -- node [arrow_description] {rgb\_analysis.txt} coordinate[pos=0.85] (water_rgb_out) (report.north);
        \draw [arrow] (waterpreprocessing) -- node [arrow_description] {water\_preprocessed.png} (wateranalysis);
        \draw [arrow] (wateranalysis) -- node [arrow_description] {water\_analysis.txt} (waterrgb);
        \draw [arrow] (rgbanalysis.south) -- ++(0.5,0) -- node [arrow_description] {rgb.png} (waterrgb.west);
        \draw [soft_arrow] (waterrgb) |- node [arrow_description, near end] {\textit{(modification)}} (water_rgb_out);
        \draw [arrow] (moistureanalysis) |- node [arrow_description, near start] {moisture\_analysis.txt} (report);

        \draw [arrow] (report) -- (llm);
    \end{tikzpicture}
    }
    \caption{Overview of all EcoScapes modules and the final use as a knowledge base. The second line in each module is the processing method or LLM used in this step.}
    \label{fig:ecoscapes_modules_overview}
\end{figure}

EcoScapes is a modular framework tailored to automate the generation of localized climate adaptation reports.
Its architecture integrates high-resolution satellite data, multimodal LLMs (MLLM), and a structured internal pipeline that adapts to varied urban settings.
An overview of the system's components and data flow is depicted in Figure~\ref{fig:ecoscapes_modules_overview}.

\subsection{Loading Satellite Images}
The workflow begins with the conversion of a user-specified town or city name into geographic coordinates using the Nominatim geocoding service~\cite{nominatim}.
A bounding box of 5~km around the central coordinates is used as the spatial extent for satellite data retrieval.
With this information, the \texttt{rgb.png}, \texttt{moisture.png} and \texttt{water.png} images shown in Figure \ref{fig:methodology:satellite_image_comparison} are created.

\subsection{Modular Pipelines}
EcoScapes employs a modular structure, built around the \texttt{Module} class, to facilitate customizable internal workflows.
This class employs two key attributes: Dependencies and soft dependencies, inspired by the Spigot plugin loading definitions~\cite{spigotmc_plugin_yml}.
Here, soft dependencies may fail or terminate without stopping the internal pipeline.
This provides flexibility in cases where a previous module might not be critical to the overall workflow, allowing for partial failures without disrupting the entire internal pipeline.
This structure ensures that each module in an internal pipeline is processed in the correct order, facilitating robust task execution and minimizing the risk of errors.

\subsection{Analytical Modules}

\subsubsection{RGB Analysis}
This mandatory module is the core spatial assessment tool, utilizing the 360VL model~\cite{360vlmodel}, a perception-oriented MLLM, to interpret the \texttt{rgb.png} image.
The module identifies urban features like built structures, green spaces, and road patterns.
To avoid hallucinations and maximize accuracy, the analysis is driven by a series of small, focused prompts rather than a monolithic query.
This modular prompt design, refined through empirical evaluation, ensures detailed and grounded responses.

\subsubsection{Water Pipeline}
An initial preprocessing step involving thresholding and denoising eliminates irrelevant water features like small pools from the \texttt{water.png} image.
If significant water bodies exist, the 360VL model evaluates their form and context.
The final Water RGB-Analysis step evaluates the \texttt{rgb.png} again with the extracted information about water bodies to assess proximity to infrastructure and the presence of natural buffers.

\subsubsection{Moisture Analysis}
Operating independently on the \texttt{moisture.png} image, this module leverages 360VL to identify moisture distribution anomalies associated with urban heat islands or cooler zones such as parks and rivers.

\subsection{Climate Report Generation}
The concluding module, \textit{Climate Report}, synthesizes the data from the preceding analytical modules into one textual report.
For this task, the InternLM model~\cite{cai2024internlm2technicalreport} was selected.
It was chosen for its high performance in open-weight benchmarks among models under 10 billion parameters and for its compatibility with local deployment.
The model is primed with a system prompt instructing it to behave as a climate scientist, encouraging outputs grounded in scientific reasoning and environmental relevance.
The model receives a concatenated input of all previous module outputs and generates a detailed summary describing environmental risks and current adaptation readiness.

\subsection{ChatClimate Knowledge Base}
The final step to utilize the \textit{Climate Report} output is the integration into a larger LLM as a knowledge base.

As CC~\cite{vaghefi2023chatclimate} showed superior accuracy compared to normal LLMs, we decided to recreate this system for our evaluation of EcoScapes.
For this, we used the ChatGPT~\cite{chatgpt} builder feature, adding the IPCC AR6 as a knowledge file and instructing it to behave like a scientific advisor for cities.

This system is then given the \textit{Climate Report} output as part of its prompt to assess the impact of EcoScapes.

The complete list of prompts employed across all modules is available in Appendix \ref{appendix:prompts}.

\section{Evaluation}

\subsection{Testing Approach}
Standard machine learning benchmarks fall short when evaluating systems like EcoScapes, which generate outputs that require creativity, contextual awareness, and nuanced reasoning.
Due to the inherently subjective nature of creative and analytical outputs, it is practically impossible to rely solely on standardized quantitative tests.
Instead, this study employs a qualitative approach, necessitating a manual evaluation of the generated climate adaptation strategies.

\subsection{Case Study Selection}
Two contrasting urban environments were deliberately chosen to reflect a broad range of urban planning challenges:
Roßtal, a small, resource-limited rural community, (population approximately 10,000) serves as a case study for understanding climate adaptation in rural areas with less funds and simpler infrastructure demands.
Conversely, Erlangen, a medium-sized city with more complex infrastructure, (population over 100,000) provides insights into climate adaptation challenges in larger urban settings, where infrastructure complexity, industrial activities, and higher population density require more nuanced and extensive adaptation measures. 
Due to the extensive nature of manual grading required for the comprehensive qualitative analysis, only these two case studies were evaluated, thereby limiting generalizability but effectively demonstrating proof-of-concept.

\subsection{Comparison Setup}
To rigorously assess the effectiveness of EcoScapes, its integration with CC~\cite{vaghefi2023chatclimate} was compared against an unchanged CC system.
EcoScapes-enhanced CC was additionally provided with context-specific climate reports generated by Eco\-Scapes, aiming to validate whether local insights improved the relevancy and accuracy of adaptation recommendations.

Evaluation utilized distinct criteria for EcoScapes reports and the final climate adaptation strategies produced by CC.
The EcoScapes \textit{Climate Report} outputs were assessed for correctness and depth \& coverage, with scores given as follows:\\
\begin{tcolorbox}
[title=Evaluation criteria for EcoScapes. Examples for scores are in parenthesis.,
title filled=false,
breakable,
colback=black!5,
colframe=black!90,
boxrule=0.5pt]

\textbf{Correctness}
\begin{description}[
    style=multiline,
    font=\normalfont\bfseries
]
  \item[0\,:] Massive mistakes (`cool coastal town'' for inland city with heat issues)
  \item[1\,:] Major mistakes in one part of the analysis (inland city confused for coastal one)
  \item[2\,:] Medium mistakes (described nonexistent lakes or heat islands)
  \item[3\,:] Many smaller mistakes (confused rails with roads or forests with fields)
  \item[4\,:] Few smaller mistakes (confused rails with roads or forests with fields)
  \item[5\,:] No mistakes
\end{description}

\textbf{Depth \& Coverage:}
\begin{description}[
    style=multiline,
    font=\normalfont\bfseries
]
    \item[0\,:] Skips most sections or gives too vague info (only mentions `a town with buildings'')
    \item[1\,:] Skips a major section like water and flood management when needed for the city
    \item[2\,:] Covers all important sections, but omits key details (rivers and lakes in the city for flooding)
    \item[3\,:] Covers all important sections, medium depth (heat problems without location)
    \item[4\,:] Covers all important sections, some vagueness (heat problems in industrial districts)
    \item[5\,:] Detailed coverage of all important sections
\end{description}
\end{tcolorbox}\

Meanwhile, the final strategies output by CC were assessed for usability, correctness and relevancy, with scores assigned as follows:\\

\begin{tcolorbox}[title=Evaluation criteria for the climate adaptation strategies of CC or CC + EcoScapes. Examples for the score are in parenthesis.,
breakable,
title filled=false,
colback=black!5,
colframe=black!90,
boxrule=0.5pt]
\textbf{Usability:}
\begin{description}[
    style=multiline,
    font=\normalfont\bfseries
]
    \item[0\,:]  Output is too vague or generic to be helpful (`Flood protection should be improved'')
    \item[1\,:] Output presents general suggestions with little detail (`improving green spaces'' without specific ideas)
    \item[2\,:] Output gives basic, practical suggestions but lacks creativity (recommends `building flood walls'')
    \item[3\,:] Output is detailed but lacks practical aspects (suggests ideas like rain gardens in highly urbanized space)
    \item[4\,:] Output is detailed with some alterations needed (for example, suggests multiple solutions of which some are not practical)
    \item[5\,:] Output is detailed, creative, and practical
\end{description}

\textbf{Correctness:}
\begin{description}[
    style=multiline,
    font=\normalfont\bfseries
]
    \item[0\,:] Output contains misinformation (`Climate change is not real'')
    \item[1\,:] Major factual errors (estimates completely incorrect environmental data)
    \item[2\,:] Several significant factual inaccuracies (wrong climatic trends for the area)
    \item[3\,:] Some minor inaccuracies (some misrepresentation of climate trends)
    \item[4\,:] Mostly factually correct, with only small details wrong
    \item[5\,:] Fully accurate
\end{description}

\textbf{Relevancy:}
\begin{description}[
    style=multiline,
    font=\normalfont\bfseries
]
    \item[0\,:] Entirely ignores the local context (flood protection for a desert town)
    \item[1\,:] Mostly irrelevant to the local context (flood protection for nonexistent rivers)
    \item[2\,:] Partially relevant or vague answers (`flood management should be improved'', no concrete measures)
    \item[3\,:] Generally relevant, but hard to implement(suggests extensive park areas in a highly urbanized zone)
    \item[4\,:] Mostly relevant with only minor mismatches (fits the local context but emphasizes secondary issues)
    \item[5\,:] Perfectly aligned with the local context
\end{description}
\end{tcolorbox}\

\subsection{Bias and Transparency}
Various biases were identified that we could not or only partially account for:
\begin{itemize}
\item \textbf{Environmental Bias:} Variability and limitations inherent to satellite imagery can impact accuracy, particularly due to factors such as seasonal differences, atmospheric conditions, and vegetation cover.
\item \textbf{Policy Bias:} Recommendations may inherently reflect certain policy biases or preferences embedded in the pre-training data of the LLMs used in EcoScapes or the IPCC AR6 used for CC, potentially skewing results toward more common or documented strategies.
\item \textbf{Model Bias:} Inherent limitations and biases of large language models used could affect interpretations and accuracy, potentially leading to misclassification or misrepresentation of urban features.
\item \textbf{Evaluator Bias:} Personal interpretations by the evaluator during manual grading can introduce subjective biases. 
This bias was carefully considered and partially mitigated by clearly defining evaluation criteria and maintaining transparency in scoring. 
Moreover, blinding the source of the texts was discussed, but due to differences in the resulting report style and frequent references of the EcoScapes results for the EcoScapes-enhanced CC system, this was considered impractical.
\end{itemize}

To further mitigate biases and ensure reliability, each scenario underwent multiple evaluation runs, reducing variability introduced by the model temperature.
For full transparency, all results and the full report texts can be found in the Appendix \ref{appendix:ecoscape_answers}.
\section{Results}

\subsection{EcoScapes Framework Evaluation}

\begin{figure}[htbp]
    \centering

    \begin{subfigure}[b]{0.48\textwidth}
        \centering
        \begin{tikzpicture}
            \begin{axis}[
                boxplot/draw direction=y,
                ylabel={Scores (0-5)},
                xtick={1,2},
                xticklabels={Roßtal, Erlangen},
                title={Correctness (EcoScapes)},
                ymajorgrids=true,
                ymin=0, ymax=6,
                width=\linewidth,
                height=4.5cm,
            ]
                \addplot+ [chatclimate ecoscapes] table [row sep=\\,y index=0] {
                    data,
                    4\\5\\4\\5\\5\\
                };
                \addplot+ [chatclimate ecoscapes] table [row sep=\\,y index=0] {
                    data,
                    5\\4\\5\\4\\3\\
                };
            \end{axis}
        \end{tikzpicture}

        \caption{Correctness}
        \label{fig:ecoscapes_correctness}
    \end{subfigure}
    \hfill
    \begin{subfigure}[b]{0.48\textwidth}
        \centering
        \begin{tikzpicture}
            \begin{axis}[
                boxplot/draw direction=y,
                ylabel={Scores (0-5)},
                xtick={1,2},
                xticklabels={Roßtal, Erlangen},
                title={Depth \& Coverage (EcoScapes)},
                ymajorgrids=true,
                ymin=0, ymax=6,
                width=\linewidth,
                height=4.5cm,
            ]
                \addplot+ [chatclimate ecoscapes] table [row sep=\\,y index=0] {
                    data,
                    4\\1\\4\\4\\4\\
                };
                \addplot+ [chatclimate ecoscapes] table [row sep=\\,y index=0] {
                    data,
                    2\\2\\4\\2\\2\\
                };
            \end{axis}
        \end{tikzpicture}

        \caption{Depth \& Coverage}
        \label{fig:ecoscapes_depth}
    \end{subfigure}

    \caption{EcoScapes Results for Roßtal vs.\ Erlangen}
    \label{fig:ecoscapes_standalone_horizontal}
\end{figure}
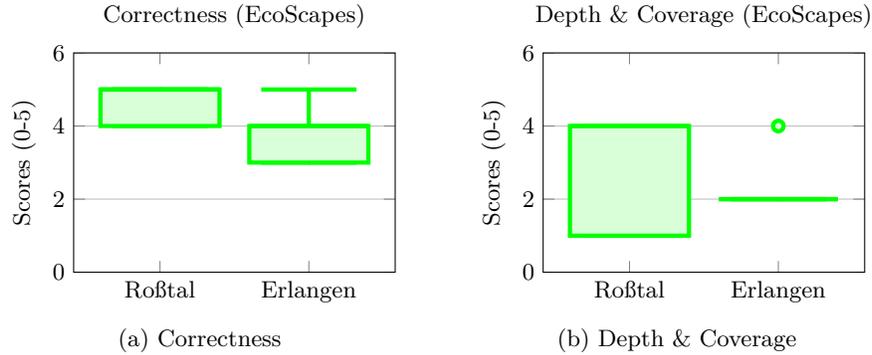

Figure~\ref{fig:ecoscapes_standalone_horizontal} provides a comparative overview of EcoScapes' performance across the two studied locations.
EcoScapes exhibited notable challenges, especially regarding correctness.
This was particularly evident in the heat analysis segment, where superficial and repetitive information overshadowed deeper analytical insights.

EcoScapes also encountered hallucinations, erroneously identifying details such as vehicles, which are unlikely discernible given the spatial resolution constraints of Sentinel-2 imagery, as illustrated in Figure~\ref{fig:methodology_limitations_resolution}.

Depth and coverage varied significantly. 
For instance, the omission of critical geographic features such as rivers in Erlangen's reports indicated inconsistency in data integration despite successful identification in other parts of the analysis.
An isolated incident in Roßtal's heat analysis generated nonsensical output, leading to a singularly poor rating; however, this occurrence was not repeated in subsequent trials.

\subsection{Climate Adaptation Strategy Generation}

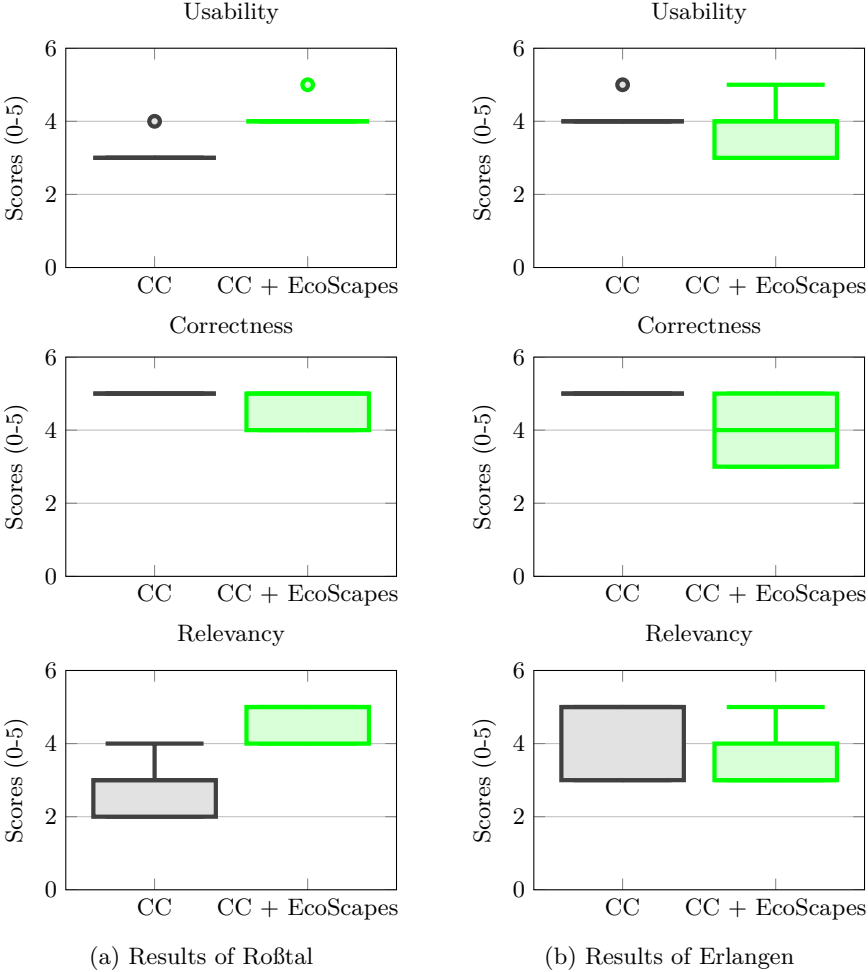
\begin{figure}[htbp]
    \centering
    
    \begin{subfigure}[b]{0.49\textwidth}
        \centering
        \begin{tikzpicture}
            \begin{axis}[
                boxplot/draw direction=y,
                ylabel={Scores (0-5)},
                xtick={1,2},
                xticklabels={CC, CC + EcoScapes},
                title={Usability},
                ymajorgrids=true,
                ymin=0, ymax=6,
                width=\linewidth,
                height=4.5cm,
            ]
                \addplot+ [chatclimate pure] table [row sep=\\,y index=0] {
                    data,
                    3\\3\\4\\3\\3\\
                };
                \addplot+ [chatclimate ecoscapes] table [row sep=\\,y index=0] {
                    data,
                    4\\4\\4\\5\\4\\
                };
            \end{axis}
        \end{tikzpicture}

        \begin{tikzpicture}
            \begin{axis}[
                boxplot/draw direction=y,
                ylabel={Scores (0-5)},
                xtick={1,2},
                xticklabels={CC, CC + EcoScapes},
                title={Correctness},
                ymajorgrids=true,
                ymin=0, ymax=6,
                width=\linewidth,
                height=4.5cm,
            ]
                \addplot+ [chatclimate pure] table [row sep=\\,y index=0] {
                    data,
                    5\\5\\5\\5\\5\\
                };
                \addplot+ [chatclimate ecoscapes] table [row sep=\\,y index=0] {
                    data,
                    5\\4\\5\\5\\5\\
                };
            \end{axis}
        \end{tikzpicture}

        \begin{tikzpicture}
            \begin{axis}[
                boxplot/draw direction=y,
                ylabel={Scores (0-5)},
                xtick={1,2},
                xticklabels={CC, CC + EcoScapes},
                title={Relevancy},
                ymajorgrids=true,
                ymin=0, ymax=6,
                width=\linewidth,
                height=4.5cm,
            ]
                \addplot+ [chatclimate pure] table [row sep=\\,y index=0] {
                    data,
                    3\\2\\4\\3\\3\\
                };
                \addplot+ [chatclimate ecoscapes] table [row sep=\\,y index=0] {
                    data,
                    5\\5\\5\\4\\5\\
                };
            \end{axis}
        \end{tikzpicture}
        
        \caption{Results of Roßtal}
        \label{fig:rosstal_left_subfig}
    \end{subfigure}
    \hfill
    \begin{subfigure}[b]{0.49\textwidth}
        \centering
        \begin{tikzpicture}
            \begin{axis}[
                boxplot/draw direction=y,
                ylabel={Scores (0-5)},
                xtick={1,2},
                xticklabels={CC, CC + EcoScapes},
                title={Usability},
                ymajorgrids=true,
                ymin=0, ymax=6,
                width=\linewidth,
                height=4.5cm,
            ]
                \addplot+ [chatclimate pure] table [row sep=\\,y index=0] {
                    data,
                    4\\4\\5\\4\\4\\
                };
                \addplot+ [chatclimate ecoscapes] table [row sep=\\,y index=0] {
                    data,
                    5\\5\\4\\3\\4\\
                };
            \end{axis}
        \end{tikzpicture}

        \begin{tikzpicture}
            \begin{axis}[
                boxplot/draw direction=y,
                ylabel={Scores (0-5)},
                xtick={1,2},
                xticklabels={CC, CC + EcoScapes},
                title={Correctness},
                ymajorgrids=true,
                ymin=0, ymax=6,
                width=\linewidth,
                height=4.5cm,
            ]
                \addplot+ [chatclimate pure] table [row sep=\\,y index=0] {
                    data,
                    5\\5\\5\\5\\5\\
                };
                \addplot+ [chatclimate ecoscapes] table [row sep=\\,y index=0] {
                    data,
                    5\\5\\4\\3\\5\\
                };
            \end{axis}
        \end{tikzpicture}

        \begin{tikzpicture}
            \begin{axis}[
                boxplot/draw direction=y,
                ylabel={Scores (0-5)},
                xtick={1,2},
                xticklabels={CC, CC + EcoScapes},
                title={Relevancy},
                ymajorgrids=true,
                ymin=0, ymax=6,
                width=\linewidth,
                height=4.5cm,
            ]
                \addplot+ [chatclimate pure] table [row sep=\\,y index=0] {
                    data,
                    3\\3\\5\\5\\5\\
                };
                \addplot+ [chatclimate ecoscapes] table [row sep=\\,y index=0] {
                    data,
                    4\\3\\4\\3\\5\\
                };
            \end{axis}
        \end{tikzpicture}

        \caption{Results of Erlangen}
        \label{fig:erlangen_right_subfig}
    \end{subfigure}

    \caption{Comparison of CC (left bars) vs. CC + EcoScapes (right bars) for (a) Roßtal and (b) Erlangen across the criteria Usability, Correctness and Relevancy}
    \label{fig:rosstal_vs_erlangen}
\end{figure}

\subsubsection{Roßtal Case Study}

Integration of EcoScapes with ChatClimate improved the relevancy and usability of climate adaptation strategies for Roßtal (Figure~\ref{fig:rosstal_left_subfig}).
The base CC system initially struggled with contextual accuracy, proposing irrelevant measures such as flood barriers unsuitable to Roßtal's geographical context.

In contrast, the EcoScapes integration provided critical local insights, substantially enhancing the tailored recommendations. 
The model adjusted its outputs to align closely with Roßtal’s resources and capacities, demonstrating superior performance in relevancy and usability than CC without the EcoScapes report.

\subsubsection{Erlangen Case Study}

Contrastingly, the performance of the integrated system for Erlangen, a medium-sized city, showed a notable decline compared to the standalone ChatClimate model (Figure~\ref{fig:erlangen_right_subfig}).
The base CC model, leveraging prior pre-trained knowledge of the larger and more well-known city, effectively referenced key local features, including the Regnitz river.

EcoScapes' inaccuracies in water-related data compromised this integration, introducing uncertainty in subsequent recommendations.
As a result, the integrated model exhibited reduced confidence and accuracy, highlighting EcoScapes’ limitations when dealing with well-documented urban environments.

\subsection{Discussion}

The evaluation illustrates that while EcoScapes notably improves local climate adaptation strategies in contexts lacking comprehensive pre-existing data, its performance is inconsistent in more well-known urban environments that likely appear in pre-training data.
Overall, EcoScapes achieved its goal to provide a benefit for smaller towns in our small case-study as a proof-of-concept, but further evaluation on a broader selection of towns and cities is required to generalize the results.
\section{Conclusion}

This paper aims to enhance the CC system by increasing the local relevance of its climate adaptation strategies through integration with EcoScapes, a system that provides insights derived from local satellite image analysis.
While a comprehensive analysis has yet to be conducted, preliminary findings suggest EcoScapes may offer particular benefits for smaller towns, where it can supplement the base model’s general knowledge.
In larger and more well-documented urban areas, however, the base model's knowledge alone may suffice or outperform the information provided by EcoScapes.

Further research is necessary to validate the limited data presented here and to refine the EcoScapes framework.
Other potential enhancements are a more intricate and exhaustive satellite image processing, more pipelines and better satellite image resolution, which may enable EcoScapes to deliver more detailed and higher-quality information to CC, thus improving the system’s overall applicability.

This paper demonstrates the feasibility of using freely available satellite imagery as a resource for enriching the CC system’s responses.
Despite its limitations, the Sentinel-2 satellite data has proven sufficient for informing climate adaptation strategies in smaller towns that, as highlighted by the IPCC AR6, often possess fewer resources and are thus more vulnerable.
The targeted local insights provided by EcoScapes represent a valuable addition to climate resilience efforts in these contexts.


\section{Accessing and Using the Copernicus Data Browser}\label{appendix:copernicus_data_browser}

The Copernicus Data Browser is a free online platform that allows users to access and download satellite imagery from the Sentinel missions. The following step-by-step guide provides instructions on how to manually obtain satellite images using the Copernicus Data Browser:

\subsection{Step-by-Step Instructions}

\begin{enumerate}
    \item \textbf{Navigate to the Copernicus Data Browser}
    \begin{itemize}
        \item Open your web browser and go to the Copernicus Data Browser:
        
        \url{https://browser.dataspace.copernicus.eu/}
    \end{itemize}
        
    \item \textbf{Set the Area of Interest}
    \begin{itemize}
        \item Use the map to navigate to your area of interest.
        \item Zoom in and out using the scroll wheel or zoom buttons.
        \item To define a specific area:
        \begin{itemize}
            \item Use the rectangle tool left of the \textbf{``Create area of interest''} icon, a pentagon, in the top right corner.
            \item Draw a rectangle or polygon over your desired area.
        \end{itemize}
    \end{itemize}

    \item \textbf{Set maximum cloud cover}
    \begin{itemize}
        \item On the left upper part of the screen, click the cloud icon and set at most 1\% cloud cover. 
    \end{itemize}

    \item \textbf{Specify the Time Frame}
    \begin{itemize}
        \item Click \textbf{``Show latest date''}.
        \item If the latest date is now too far back in time, go to the previous step and increase the maximum cloud cover.
    \end{itemize}
    
    \item \textbf{Select Layer}
    \begin{itemize}
        \item The lower left part of the screen contains the layer selection.
        
        \item \textbf{``True color''} results in the \texttt{rgb.png} image in figure \ref{fig:methodology:satellite_image_comparison:rgb}
        \item \textbf{``Moisture index''} results in the \texttt{moisture.png} image in figure \ref{fig:methodology:satellite_image_comparison:moisture}
        \item For the \texttt{water.png} image in figure \ref{fig:methodology:satellite_image_comparison:water}, you need to select \textbf{``Custom''}, then click on \textbf{``Custom script''} and finally paste in this code:

\begin{lstlisting}[language=Java,breaklines=true]
//VERSION=3
const colorRamp = [[0,0xffffff],[1,0x000000]]

let viz = new ColorRampVisualizer(colorRamp);

function setup() {
  return {
    input: ["B03","B08", "dataMask"],
    output: [
      { id:"default", bands: 4 },
      { id: "index", bands: 1, sampleType: 'FLOAT32' }
    ]
  };
}

function evaluatePixel(samples) {
  let index = (samples.B03 - samples.B08) / (samples.B03 + samples.B08);
  const minIndex = 0;
  const maxIndex = 1;
  let visVal = null;

  if (index > maxIndex || index < minIndex) {
    visVal = [0, 0, 0, 0];
  } else {
    visVal = [...viz.process(index), samples.dataMask];
  }

  // The library for tiffs only works well if there is one channel returned.
  // So here we encode "no data" as NaN and ignore NaNs on the frontend.
  const indexVal = samples.dataMask === 1 ? index : NaN;

  return { default: visVal, index: [indexVal] };
}
\end{lstlisting}

    Note that the water image, in particular, might differ from the automatically generated one. 
    Indeed, the Data Explorer's images usually indicate more water.
    \end{itemize}
    
    \item \textbf{Download}
    \begin{itemize}
        \item On the right side, click the \textbf{``Download image''} button.
        \item Ensure \textbf{``Crop to AOI''} is selected (at time of writing it is selected by default).
        \item Leave the \textbf{``Description''} field empty.
        \item Change the image format to \textbf{``PNG (no georeference)''}
        \item Click on \textbf{``Download''}
    \end{itemize}
    
    \item \textbf{Edit image}
    \begin{itemize}
        \item Rename the image to \texttt{rgb.png}, \texttt{moisture.png} or \texttt{water.png}, depending on the selected layer.
        \item Crop the image in your favorite image editor to remove the branding and scales.
        \item Rescale the image proportionally, ensuring the largest side is 1024 pixels wide.
    \end{itemize}
    
    \item \textbf{Input to Ecoscapes}
    \begin{itemize}
        \item In the same folder as the \texttt{main.py} file of EcoScapes, create a new folder called \texttt{satellite\_data} (case sensitive).
        \item In this new folder, create another folder with the name of your requested city.
        \item Move your downloaded and edited image to the newly created city folder.
    \end{itemize}
    
    \item \textbf{Repeat as Necessary}
    \begin{itemize}
        \item Repeat until all three layers are downloaded.
        \item Ensure that the selected area does not differ significantly between the different layers.
    \end{itemize}
\end{enumerate}

\clearpage
\section{Prompts}\label{appendix:prompts}

\begin{table}[htb]
\caption{All prompts of the RGB-Analysis module}
\label{tab:rgb_analysis_prompts}
\centering
\begin{tabularx}{\textwidth}{cX}
\toprule
\textbf{\#} & \textbf{Prompt} \\
\midrule
1  & Please approximate the size and population of the town or city. \\
2  & Is the town or city layout grid-like, circular, or irregular? \\
3  & Where in the picture is the city center located? Please describe the location and the distance from the center to the edges of the town. \\
4  & Are there any forests or parks in the image? If yes, please describe their location and size. \\
5  & Are there any railway lines or stations in the image? If yes, please describe their location and connections. \\
6  & Are there any major bridges or tunnels in the image? If yes, please describe their location and connections. \\
7  & How are the buildings distributed in the town? Are there any residential, commercial, industrial, or mixed zones in the image? Please describe their location and density, if it is visible from the picture. Otherwise, say that you are unable to identify specific zones. \\
8  & What is the approximate distribution and density of buildings in the town or city? Are there clusters of high-density areas? \\
9  & Can you identify large open spaces within the urban area? Describe their locations and approximate sizes. \\
10 & Are there distinct patterns in the road network? Describe the layout and any major intersections. \\
11 & Is there a visible distinction between urban and rural areas? Describe the transition zones and their locations. \\
12 & Can you identify any major landmarks or large structures that stand out? Describe their locations relative to the city center. \\
13 & Are there large expanses of greenery or parkland visible? Describe their locations and approximate sizes. \\
14 & Can you see any significant geographical features, such as hills or valleys, in or around the city? Describe their locations. \\
\bottomrule
\end{tabularx}
\end{table}
\begin{table}[htb]
\caption{Prompts for MoistureAnalysis}
\label{tab:moisture_analysis_prompts}
\centering
\begin{tabularx}{\textwidth}{cX}
\toprule
\textbf{\#} & \textbf{Prompt} \\
\midrule
1 & List specific spots with the highest heat levels (red areas). \\
2 & List specific spots with the lowest heat levels (blue areas). \\
3 & Compare the heat levels between different sectors (e.g., urban vs. rural). \\
4 & Identify patterns or trends in heat distribution (e.g., is there a gradient?). \\
5 & Analyze potential reasons for red high heat spots (e.g., industrial areas, lack of vegetation). \\
6 & Analyze potential reasons for blue low heat spots (e.g., water bodies, dense vegetation). \\
7 & Assess the implications of heat distribution on urban infrastructure. \\
\bottomrule
\end{tabularx}
\end{table}

\begin{table}[htb]
\caption{Prompts for WaterAnalysis module. System prompts give basic instructions to the model and are retained across all User prompts, which start individual chats.}
\label{tab:water_analysis_prompts}
\centering
\begin{tabularx}{\textwidth}{cX}
\toprule
\textbf{\#} & \textbf{Prompt} \\
\midrule
1 & \textbf{System:} The map shows water as white and land as black. A river is a very long, connected, white area. A lake is a large, circular, white area. \\
2 & \textbf{User:} Is the map depicting a lake? If it does not, please say so. \\
3 & \textbf{User:} Is the map depicting a river? If it does not, please say so. \\
4 & \textbf{User:} Is the map depicting a part of the coast? If it does not, please say so. \\
\bottomrule
\end{tabularx}
\end{table}

\begin{table}[htb]
\caption{Prompts for ClimateReport}
\label{tab:climate_report_prompts}
\centering
\begin{tabularx}{\textwidth}{cX}
\toprule
\textbf{\#} & \textbf{Prompt} \\
\midrule
1 & \textbf{System:} You are a climate scientist with a focus on climate adaptation. You will be given tasks that will result in a report to analyze the current state of climate adaptation in a city or town. Answer accurately, informatively, and in a neutral way that aligns with the scientific consensus. \\
2 & \textbf{User:} You will be given a report of the RGB satellite image of the city or town \texttt{<location>} and its surrounding area and a description of moisture anomalies of the town and its surroundings taken on a sunny day. Please use this information to write a report on the current state of climate adaptation of the town. Only focus on the current situation and do not make any predictions. \\
3 & \textbf{User:} The RGB satellite image description: \texttt{rgb\_analysis.txt} The moisture map description: \texttt{moisture\_analysis.txt} \\
\bottomrule
\end{tabularx}
\end{table}

\clearpage
\section{Evaluation - EcoScape Outputs}\label{appendix:ecoscape_answers}

\subsection{Run 1 - Roßtal}

\subsubsection{Unchanged ChatClimate}

To develop a detailed climate adaptation strategy for Rosstal, it is essential to first understand the region's specific climate risks, projected impacts, and socio-economic structure. While general principles of climate adaptation apply, strategies must be tailored to local conditions. The adaptation strategy for Rosstal will focus on urban greening, energy efficiency, water management, and resilient infrastructure.

1. Climate Projections and Risks for Rosstal

    Temperature Increase: Rosstal, like other parts of Bavaria, will experience hotter summers with more frequent and intense heatwaves. The average global temperature is projected to increase by 1.5°C by the early 2030s, with potentially greater impacts at the regional level.
    Precipitation Patterns: Changing precipitation patterns may lead to more erratic weather, with periods of drought interspersed with intense rainfall, increasing the risk of both droughts and localized flooding.
    Extreme Weather Events: These shifts in weather patterns may intensify storm events and heat-related health risks, particularly affecting vulnerable populations like the elderly.

2. Urban Greening and Nature-Based Solutions

    Increase Urban Green Spaces: Integrate green roofs, parks, and urban forests to reduce heat island effects and enhance air quality. Rosstal can benefit from the ecosystem-based adaptation approaches suggested by the IPCC, such as urban greening and wetland restoration, which have been shown to effectively reduce flood risks and manage heat.
    Water Retention Gardens: These gardens can help manage stormwater, reduce runoff, and mitigate flooding risks. They also contribute to biodiversity and the cooling of urban areas.
    Reforestation and Agroforestry: Enhancing tree cover both within and around the town can serve as carbon sinks, reduce temperature, and help manage excess rainwater through natural filtration.

3. Energy Efficiency and Renewable Energy

    Building Retrofitting: Encourage retrofitting existing buildings with energy-efficient technologies such as better insulation, smart lighting, and efficient cooling systems to prepare for hotter summers. This reduces both energy demand and carbon emissions.
    District Energy Systems: Implement district heating or cooling systems, leveraging renewable energy sources such as solar or geothermal to reduce reliance on fossil fuels.
    Solar and Wind Energy: Invest in local renewable energy infrastructure, including solar panels on buildings or community wind projects, to ensure energy resilience in the face of climate-induced disruptions to the grid.

4. Water Management

    Drought-Resistant Water Systems: With the likelihood of increasing droughts, investing in water-saving technologies, such as rainwater harvesting and greywater recycling systems, will be crucial to ensure a sustainable water supply during dry periods.
    Flood-Resilient Infrastructure: Strengthen water management systems to prevent urban flooding through improved drainage systems and permeable surfaces in urban planning, allowing for better water absorption during heavy rainfall.

5. Resilient Infrastructure

    Heat-Resilient Infrastructure: Infrastructure should be designed to cope with extreme heat. For example, materials for roads and buildings can be chosen for their ability to withstand high temperatures and prevent buckling or damage.
    Flood-Resilient Urban Design: Implement flood barriers where necessary, and elevate critical infrastructure (such as power stations, hospitals, and roads) to prevent damage from possible flash floods.

6. Community and Governance Involvement

    Early Warning Systems: Establish community-based early warning systems for extreme weather events, ensuring timely communication of risks such as heatwaves or storms to vulnerable populations.
    Public Engagement and Education: Raise awareness through public engagement campaigns on energy efficiency, water conservation, and climate risks. Initiatives like climate education in schools or community workshops on sustainable practices will help build a resilient society.
    Equity and Inclusiveness: Focus on protecting vulnerable populations (elderly, children, low-income households) from climate impacts through targeted financial and social safety net programs.

7. Cross-Sectoral Integration

    Urban Planning Integration: Integrate climate adaptation strategies into Rosstal's urban development plans, ensuring that new developments comply with sustainable building codes and incorporate green spaces, flood resilience, and renewable energy sources.
    Collaboration with Neighboring Municipalities: Climate risks like droughts and storms often extend beyond local borders. Collaborative planning with neighboring towns can optimize regional adaptation strategies, such as shared renewable energy infrastructure or coordinated emergency responses.

8. Monitoring and Adjustments

    Monitoring Climate Impacts: Establish a local climate task force to regularly assess the effectiveness of adaptation measures and adjust them as new climate projections emerge.
    Funding for Research and Adaptation: Secure funding for local adaptation initiatives through national and EU climate funds, and ensure that financial resources are allocated to both mitigation and adaptation efforts.

Conclusion

By integrating nature-based solutions, energy efficiency, resilient infrastructure, and strong community involvement, Rosstal can develop a robust climate adaptation strategy. This approach ensures the town not only adapts to the immediate climate risks but also builds a sustainable, resilient future for its residents.

\subsubsection{EcoScapes}

**Climate Adaptation Report: The Town of Rosstal and Its Surroundings**

**Executive Summary:**
This report analyzes the current state of climate adaptation in the town of Rosstal and its surrounding area. The analysis is based on the RGB satellite image of the town and its moisture anomalies taken on a sunny day. The report focuses on the town's layout, infrastructure, green spaces, and moisture patterns to assess its climate adaptation strategies and vulnerabilities.

**1. Town Layout and Infrastructure:**
Rosstal is characterized by a relatively small size, with a population estimated between 10,000 to 50,000 people. The town's layout is irregular, featuring a mix of buildings and green spaces. The city center is situated in the middle of the image, surrounded by expansive green fields. The town's size and open spaces contribute to a sense of tranquility and openness, distinguishing it from densely populated urban centers.

The town boasts a well-developed transportation network, including railway lines and stations that run through the center, facilitating easy access to urban areas. Two major bridges connect different parts of the city, enhancing connectivity and mobility.

**2. Green Spaces and Urban Aesthetics:**
Rosstal is enriched with several forests and parks, scattered throughout the city. These green spaces enhance the town's aesthetic and livability, providing residents with recreational areas and contributing to the overall environmental quality.

**3. Public Transportation:**
The presence of railway lines and stations indicates a robust public transportation system, reducing traffic congestion and promoting sustainable commuting options. This infrastructure is vital for climate adaptation by reducing reliance on private vehicles, which contribute to greenhouse gas emissions.

**4. Climate Vulnerabilities and Adaptation:**
The moisture anomalies map, though not provided in detail, suggests variations in temperature across different sectors of the town. Urban areas, characterized by higher heat levels, may experience increased energy demand for cooling systems, affecting energy consumption and environmental impact.

**5. Heat Island Effect:**
The heat distribution patterns, indicated by the blue and orange color scheme in the heat map, highlight areas with lower and higher heat levels. These variations can influence urban planning and infrastructure development, necessitating the integration of heat-resistant materials and natural ventilation in buildings. Additionally, the heat island effect can affect the performance and lifespan of transportation infrastructure, necessitating regular maintenance and repairs.

**Conclusion:**
Rosstal's current climate adaptation strategies reflect a balance between urban development and environmental preservation. The town's layout, transportation infrastructure, and green spaces offer a framework for sustainable living. However, challenges such as heat distribution and the heat island effect require ongoing attention and adaptation measures to ensure the town's resilience and sustainability in the face of climate change.

**Recommendations:**
1. **Enhance Green Spaces:** Invest in the maintenance and expansion of green spaces to mitigate the heat island effect and improve air quality.
2. **Energy Efficiency:** Promote energy-efficient cooling systems and improve insulation in buildings to reduce energy consumption and costs.
3. **Sustainable Transportation:** Continue to develop and maintain efficient public transportation systems to reduce reliance on private vehicles.
4. **Climate Education:** Implement community education programs to raise awareness about climate change and adaptation strategies.

By addressing these recommendations, Rosstal can strengthen its climate adaptation efforts and foster a more resilient and sustainable urban environment.

\subsubsection{EcoScapes + ChatClimate}

Based on the information provided in the Climate Adaptation Report for the town of Rosstal and surrounding areas, I will design a detailed climate adaptation strategy that incorporates the latest climate adaptation and mitigation research, referencing the Intergovernmental Panel on Climate Change (IPCC) Synthesis Report (2023) and tailored to the town's specific characteristics.

1. Overview of Rosstal's Climate Adaptation Needs
Town Characteristics:

    Population: 10,000 to 50,000 residents
    Layout: Irregular with a mix of urban and green spaces
    Public Infrastructure: Railways, major bridges, and developed transportation networks
    Green Spaces: Forests, parks, and scattered green areas
    Climate Vulnerabilities: Urban heat islands, moisture anomalies, and heat distribution imbalances

Primary Climate Challenges:

    Urban heat islands, exacerbated by heatwaves and higher summer temperatures
    Increased energy consumption for cooling due to rising temperatures
    Variability in moisture patterns, suggesting a risk of droughts or uneven water availability
    Risk of transportation and infrastructure damage from heat extremes

2. Climate Adaptation Strategy for Rosstal
2.1. Urban Greening and Heat Island Mitigation

Given Rosstal’s susceptibility to urban heat islands, increasing green space is a priority. Urban greening can reduce temperatures by providing shade and facilitating cooling through evapotranspiration. According to the IPCC's findings, urban green infrastructure is one of the most cost-effective and sustainable measures to combat heat stress.

Actions:

    Green Roofs and Walls: Expand incentives for private and public buildings to implement green roofs and vertical gardens. These measures reduce heat absorption and enhance urban cooling.
    Urban Tree Planting: Increase tree cover, especially in the town center, which experiences higher temperatures. Prioritize native and drought-tolerant species to maintain resilience.
    Park Expansion: Enhance and expand existing parks, focusing on areas where the heat island effect is most pronounced. Parks can act as cool zones during heatwaves.

2.2. Energy Efficiency and Building Retrofits

Heatwaves will increase energy demands for cooling, potentially overwhelming local energy infrastructure and contributing to higher emissions. As the IPCC recommends, energy efficiency in building design is key to reducing both emissions and energy demand.

Actions:

    Building Retrofitting: Promote insulation upgrades and shading in both public and private buildings to enhance energy efficiency. Improved insulation reduces the need for air conditioning.
    Passive Cooling Architecture: Encourage the use of passive cooling techniques in new construction, such as cross-ventilation, reflective roofs, and natural building materials that reduce heat absorption.
    Energy-Efficient Appliances: Launch campaigns promoting energy-efficient cooling systems (such as smart thermostats and heat pumps) and offer subsidies for energy-efficient appliances.

2.3. Sustainable Water Management

The presence of moisture anomalies and possible variability in rainfall points to the need for more sustainable water management, which aligns with the IPCC’s recommendations on addressing water scarcity and variability.

Actions:

    Rainwater Harvesting: Encourage the installation of rainwater harvesting systems in residential and commercial buildings, providing a supplementary water source during dry spells.
    Permeable Surfaces: Implement permeable materials in urban design, such as roads and walkways, to allow water to percolate back into the ground, reducing runoff and aiding groundwater replenishment.
    Drought-Resilient Landscaping: Foster the use of native, drought-resistant plants in public and private landscaping to reduce water use while maintaining aesthetic quality and ecosystem services.

2.4. Resilient Infrastructure and Transportation

Infrastructure, particularly transportation networks like railways and bridges, is at risk from increased heat. According to the IPCC, resilient infrastructure is essential for maintaining functionality during extreme weather events.

Actions:

    Heat-Resilient Materials: When maintaining and upgrading infrastructure (especially bridges and roads), prioritize heat-resistant materials that can withstand higher temperatures without deteriorating.
    Shading and Cooling at Transport Hubs: Install shading structures at train stations and other transport hubs to reduce heat stress on commuters. Consider solar-powered cooling options for high-traffic areas.
    Public Transport Electrification: Transition public buses and other municipal transport vehicles to electric options, reducing local emissions and dependence on fossil fuels.

3. Long-Term Adaptation and Community Resilience
3.1. Education and Community Involvement

    Climate Literacy Programs: Establish community outreach programs to raise awareness of climate change impacts and teach residents about energy efficiency, water conservation, and heat protection measures.
    Public Engagement in Urban Planning: Involve local communities in the planning and expansion of green spaces and water-saving initiatives to ensure long-term sustainability and support for adaptation efforts.

3.2. Emergency Preparedness

    Early Warning Systems: Implement real-time heat and drought warning systems to notify residents of extreme weather events, enabling them to take preventive measures.
    Cooling Centers: Designate public cooling centers equipped with energy-efficient systems to provide respite for vulnerable populations during heatwaves.

4. Financial and Governance Mechanisms
4.1. Funding for Adaptation

Securing funds for adaptation projects is essential, and IPCC emphasizes the role of public and private finance. Rosstal should seek partnerships with both government and private sectors to access climate finance.

Actions:

    Adaptation Grants and Incentives: Apply for regional, national, or EU-level adaptation funding to support infrastructure upgrades and green space projects.
    Public-Private Partnerships: Foster collaboration between the municipality and private businesses to co-fund energy efficiency and urban greening projects.

4.2. Monitoring and Evaluation

Develop a monitoring system to track the effectiveness of adaptation measures over time, making adjustments as necessary based on changing climate conditions and urban needs.

\subsection{Run 1 - Erlangen}

\subsubsection{Unchanged ChatClimate}

To develop a comprehensive climate adaptation strategy for Erlangen, a city in Bavaria, Germany, it is important to consider local climatic conditions, predicted climate change impacts, and the city’s geographical and socioeconomic context. Based on current climate models, Erlangen is expected to experience more frequent and intense heatwaves, droughts, changes in precipitation patterns, and increased risk of biodiversity loss due to changing ecosystems. Below is a detailed strategy that includes climate adaptation measures across different sectors:

1. Urban Greening and Ecosystem-Based Adaptation

    Urban Forests and Parks: Increase the city's green spaces to mitigate urban heat island effects, improve air quality, and provide shade during extreme heat events. Green corridors connecting parks could help maintain local biodiversity and support pollinators.
    Green Roofs and Walls: Encourage the use of green roofs and vertical gardens on public and private buildings. This reduces building cooling demands, improves stormwater management, and enhances urban biodiversity.
    River and Wetland Restoration: Restoration of the nearby Regnitz River and its surrounding wetlands would help mitigate flood risks and increase local biodiversity, while providing recreational space for residents.
    Sustainable Agriculture and Urban Gardens: Promote urban farming and community gardens to increase local food production, which can reduce food insecurity and offer social co-benefits such as community engagement and resilience building.

2. Water Management and Resilience to Droughts

    Stormwater Management Systems: Improve stormwater management to handle more frequent heavy rainfall events, which could overwhelm existing systems. Implement permeable pavements and rainwater harvesting systems to collect and reuse water.
    Drought-Resistant Landscaping: Encourage the use of native and drought-resistant plants in public spaces and private gardens to reduce water consumption and improve resilience to dry conditions.
    Water Conservation: Develop public campaigns and offer incentives for water-efficient appliances and behavior. Introduce water pricing that reflects scarcity to encourage conservation.

3. Heatwave Management and Public Health

    Cooling Centers and Public Spaces: Set up designated cooling centers, especially for vulnerable populations such as the elderly. These could be retrofitted public buildings or schools with enhanced cooling systems powered by renewable energy.
    Heat Action Plans: Establish a city-wide early warning system and heat action plan. This would include public awareness campaigns, advising residents on how to stay cool and hydrated, and preparing emergency services for increased health incidents.
    Urban Planning: New developments should be planned with climate resilience in mind. Streets should be designed to maximize airflow, and shaded areas should be increased to reduce the heat island effect.

4. Sustainable Infrastructure and Energy Efficiency

    Energy-Efficient Buildings: Promote retrofitting of existing buildings and implement strict energy efficiency standards for new constructions. Passive cooling techniques, such as natural ventilation and solar shading, should be encouraged.
    District Cooling Systems: Develop district cooling systems, which are more energy-efficient than individual air conditioning systems, to manage rising temperatures.
    Renewable Energy Expansion: Increase investments in renewable energy sources like solar and wind to power the city. Erlangen could also explore the potential for geothermal energy, given Bavaria's geological potential.

5. Flood and Storm Resilience

    Floodplain Protection: Preserve and restore natural floodplains along the Regnitz River to provide natural buffers against flooding. Floodplains can absorb excess rainwater, reducing the severity of floods.
    Resilient Infrastructure: Ensure that critical infrastructure, such as hospitals, emergency services, and power stations, are protected from flooding and equipped with backup power sources.
    Early Warning Systems: Enhance early warning systems for extreme weather events such as floods and storms, ensuring residents are adequately informed and prepared.

6. Transportation and Mobility

    Green Mobility: Enhance public transport and promote electric vehicles (EVs) and bicycles as part of a low-carbon mobility strategy. Establish more EV charging stations powered by renewable energy.
    Climate-Resilient Transport Networks: Upgrade road and rail infrastructure to withstand extreme weather, including heatwaves, floods, and storms. Use heat-resistant materials and elevate transport routes in flood-prone areas.

7. Biodiversity and Ecosystem Services

    Biodiversity Corridors: Create biodiversity corridors that allow species to migrate in response to changing climates. This will help maintain biodiversity and allow ecosystems to adapt.
    Invasive Species Management: Implement programs to monitor and control invasive species that may thrive under changing climate conditions, which could disrupt local ecosystems.

8. Public Awareness and Community Engagement

    Education and Engagement: Launch public awareness campaigns on climate change impacts and individual actions that can help reduce risks. This includes water conservation, energy efficiency, and community gardening.
    Citizen Science Initiatives: Encourage local communities to participate in citizen science projects that track environmental changes, such as biodiversity monitoring or air quality assessments.

9. Climate-Smart Governance and Policy

    Integrated Climate Planning: Incorporate climate adaptation into city planning processes. This would include updating zoning laws and building codes to reflect climate risks, such as extreme heat and floods.
    Collaboration with Local Institutions: Partner with the University of Erlangen-Nuremberg and local research institutions to develop innovative solutions for climate adaptation and sustainable development.
    Funding and Finance: Seek climate adaptation funding from regional, national, and European sources, such as the European Union’s Horizon Europe program, which supports climate resilience projects.

10. Monitoring and Evaluation

    Climate Impact Assessments: Regularly assess the effectiveness of climate adaptation measures, ensuring that strategies remain flexible and responsive to new scientific insights and climate predictions.
    Data-Driven Decision Making: Use data from climate models and urban monitoring systems to inform policy decisions and prioritize interventions in high-risk areas.

Conclusion

This climate adaptation strategy for Erlangen focuses on enhancing the city's resilience to heatwaves, floods, and droughts, while fostering sustainability and maintaining biodiversity. These measures would help protect vulnerable populations, reduce economic losses, and ensure a healthy, resilient environment.

\subsubsection{EcoScapes}

Title: Climate Adaptation and Urban Heat Management in Erlangen: A Report

Introduction:
Erlangen, a medium to large-sized city in Germany, showcases a blend of urban development and natural landscapes. The city's layout is characterized by a grid-like pattern, with a mix of residential, commercial, and industrial zones. The city center is surrounded by various buildings and infrastructure, extending to the edges of the town. Erlangen is well-connected through a network of roads, highways, railway lines, bridges, and tunnels. This report aims to analyze the current state of climate adaptation in Erlangen, focusing on urban heat management and its implications for urban infrastructure.

Urban Heat Management:
The city of Erlangen experiences varying heat levels, influenced by its urban layout and infrastructure. The thermal map of Erlangen, as depicted in the RGB satellite image, highlights areas of high and low heat intensity. The red areas, representing higher heat levels, are concentrated in the middle and upper parts of the city, potentially due to industrial zones, buildings, and infrastructure that generate heat. The blue areas, indicating lower heat levels, are scattered throughout the cityscape, suggesting that heat is being moderated in these regions.

Implications for Urban Infrastructure:
The implications of heat distribution on urban infrastructure are significant. High temperatures can lead to increased energy consumption, strain on the power grid, and potential overheating of buildings and infrastructure. This can result in higher maintenance costs, reduced lifespan of buildings, and increased risk of damage or failure. To address these challenges, Erlangen can implement heat island mitigation measures, such as the integration of green spaces, reflective surfaces, and urban planning that considers the heat island effect.

Green Spaces and Vegetation:
Green spaces and vegetation play a crucial role in urban heat management. Erlangen's surroundings are characterized by a large forest, which provides a natural environment for the urban area. The presence of forests and parks within the city contributes to the overall aesthetic and ecological balance. These green spaces help to absorb and reflect sunlight, reducing the urban heat island effect and providing a cooler environment for residents.

Public Transportation and Infrastructure:
Erlangen's well-developed public transportation system, including railway lines and stations, facilitates efficient movement of people and goods. The presence of major bridges and tunnels enhances the city's connectivity and functionality. These infrastructure elements contribute to the overall urban landscape and support the efficient management of urban heat.

Conclusion:
The current state of climate adaptation in Erlangen is characterized by a mix of urban development and natural landscapes. The city's grid-like layout, dense population, and diverse economic activities present unique challenges in managing urban heat. Erlangen's integration of green spaces, public transportation, and smart technologies can help mitigate the impacts of heat distribution on urban infrastructure. Continued investment in climate adaptation strategies will be essential for ensuring the sustainability and resilience of Erlangen in the face of changing climate conditions.

\subsubsection{EcoScapes + ChatClimate}

Detailed Climate Adaptation Strategy for Erlangen

Introduction: Erlangen, a medium-sized city in Germany, faces increasing climate risks, particularly from urban heat and changes in weather patterns. Like many urban areas, it is exposed to the urban heat island (UHI) effect, exacerbated by infrastructure and industrial zones. The city's adaptation strategy must address these challenges while leveraging existing assets such as public transport, green spaces, and its natural surroundings.

This strategy aims to align with international climate adaptation recommendations, such as those from the IPCC, which emphasize the need for local, context-specific actions that integrate climate resilience with sustainable urban development.

1. Heat Mitigation Strategies:

The thermal map of Erlangen highlights the prevalence of UHI effects, particularly in industrial and densely populated areas. The city must take proactive steps to mitigate heat through:

1.1. Expanding Green Spaces and Vegetation:

    Urban Greening: Develop additional green spaces within the city, focusing on areas identified as heat hotspots. This should include parks, rooftop gardens, and green corridors. The forested areas around Erlangen provide a model for integrating green spaces with urban development, creating cooler microclimates through shading and evapotranspiration.
    Tree Canopy Coverage: Increase tree planting along streets, in public squares, and around buildings. Trees provide significant cooling through shade and transpiration, reducing surface and air temperatures. Native species should be prioritized to ensure resilience to local climatic conditions.

1.2. Implementing Reflective and Permeable Surfaces:

    Cool Roofs and Pavements: Encourage the use of high-albedo (reflective) materials for buildings and pavements to reduce heat absorption. Erlangen can implement incentive programs to retrofit existing buildings with cool roofs, particularly in commercial and industrial zones. Using light-colored, permeable surfaces for pavements and roads will also help reduce heat buildup.

1.3. Water-Based Cooling Strategies:

    Urban Water Bodies and Fountains: Introduce or enhance water bodies and fountains in public spaces to promote evaporative cooling. Water features can act as heat sinks, cooling surrounding areas, and contributing to overall thermal comfort in the city.

2. Energy Efficiency and Resilient Infrastructure:

High temperatures are linked to increased energy consumption, particularly for cooling. This can strain the power grid, leading to potential blackouts and increased energy costs. Erlangen can tackle these challenges with energy efficiency measures:

2.1. Building Energy Efficiency:

    Energy-Efficient Retrofitting: Promote the retrofitting of buildings with better insulation, energy-efficient windows, and cooling systems. Incentives could be offered to homeowners and businesses to install solar panels, reducing reliance on conventional energy sources.
    Passive Cooling Design: Encourage the construction of new buildings with passive cooling designs, incorporating natural ventilation, shading devices, and reflective materials.

2.2. Smart Grid Development:

    Digital Energy Management: Implement smart grid technologies to optimize energy use across the city, especially during heat waves. Digital energy meters and real-time data management systems can help prevent grid overloads by optimizing energy consumption and storage during peak times.

3. Sustainable Transportation Solutions:

Erlangen already has a strong public transportation system that can be further enhanced to contribute to climate adaptation goals:

3.1. Expanding Public Transport Infrastructure:

    Electrification of Public Transport: Transition buses and municipal vehicles to electric models, reducing emissions and noise pollution while also making the city more climate-resilient.
    Bike and Pedestrian-Friendly Infrastructure: Invest in bicycle lanes, pedestrian zones, and shared mobility systems (e.g., e-bikes, scooters) to reduce dependence on private cars, further decreasing urban heat and air pollution.

3.2. Mobility as a Service (MaaS):

    Integrated Mobility Platforms: Develop an integrated digital platform that offers real-time data on public transport, traffic, and air quality. This will help residents make informed decisions about the most efficient and least polluting mode of transport.

4. Urban Planning and Governance:

To ensure a coordinated and sustainable approach, governance structures must be strengthened to integrate climate resilience into all aspects of urban planning.

4.1. Climate-Sensitive Zoning and Building Codes:

    Zoning Regulations: Update urban planning codes to enforce climate adaptation measures, including mandatory green roofs, tree planting, and restrictions on building materials that exacerbate heat. New developments must meet strict energy and heat management criteria.
    Building Resilience Audits: Regular audits should be conducted on critical infrastructure (bridges, tunnels, etc.) to assess vulnerability to extreme heat and other climate impacts. These audits will inform necessary retrofitting or reinforcement efforts.

4.2. Public-Private Collaboration:

    Private Sector Engagement: Foster collaborations with local industries and businesses to co-invest in resilience measures, such as green roofs and energy-efficient retrofits. Public-private partnerships can also be leveraged to fund large-scale infrastructure upgrades.

5. Community Engagement and Education:

Raising public awareness and involving citizens in adaptation efforts are essential for long-term success.

5.1. Climate Literacy Programs:

    Workshops and Information Campaigns: Launch campaigns to educate residents on how they can reduce their heat footprint, conserve energy, and adapt to rising temperatures. Workshops should focus on practical actions, such as retrofitting homes for energy efficiency and the benefits of urban greening.

5.2. Community-Led Greening Projects:

    Neighborhood Initiatives: Support community-driven projects that promote local greening, such as urban gardens, tree planting, and the maintenance of small parks. These initiatives can enhance social cohesion while contributing to climate adaptation.

6. Monitoring and Evaluation:

Erlangen must establish a robust system to monitor and evaluate the progress of its climate adaptation strategies:

6.1. Climate Data Collection:

    Urban Climate Monitoring Network: Install sensors throughout the city to measure temperature, humidity, and energy consumption in real-time. This data will allow for the continuous assessment of heat mitigation measures and inform future planning efforts.

6.2. Adaptation Performance Indicators:

    Key Metrics: Develop a set of performance indicators, such as reductions in peak temperature in key areas, energy savings, and the number of green spaces created or enhanced. These indicators will help the city track progress and adjust strategies as needed.

Conclusion:

Erlangen is well-positioned to enhance its climate resilience through an integrated strategy that addresses urban heat management, energy efficiency, and sustainable transportation. By leveraging green spaces, adopting advanced infrastructure solutions, and engaging the community, Erlangen can mitigate the impacts of climate change and ensure a livable urban environment in the coming decades. Continuous adaptation, informed by real-time data and community input, will be essential for maintaining the city's long-term sustainability and resilience.

\subsection{Run 2 - Roßtal}

\subsubsection{Unchanged ChatClimate}

To design a detailed climate adaptation strategy for Roßtal, it’s important to understand the town's specific vulnerabilities, local geography, and the broader climate trends. Roßtal is located in Bavaria, Germany, which is subject to central European climate conditions. The town may experience an increase in heatwaves, more extreme weather patterns (e.g., storms and heavy rainfall), and changes in local biodiversity due to warming temperatures. Based on current climate projections and adaptation practices, I will recommend a climate adaptation plan focusing on key areas: urban infrastructure, energy, water management, agriculture, and ecosystems.

1. Urban Infrastructure and Planning

Roßtal should integrate climate resilience into urban planning to manage the risks posed by heatwaves, heavy precipitation, and storms.
Key Measures:

    Urban Greening: Increase the presence of parks, street trees, and green roofs to mitigate heat islands, reduce stormwater runoff, and improve air quality. Urban greening strategies, such as planting native tree species, will improve the town’s capacity to absorb excess water during heavy rainfall and cool down public spaces during heat waves.
    Permeable Pavements and Rainwater Harvesting: Introduce permeable surfaces in public areas and roads to allow rainwater infiltration and reduce flood risks. Rainwater harvesting systems can be installed on public buildings for reuse in irrigation and non-potable uses.
    Heat-Resilient Building Codes: Modify building standards to include heat-resistant designs, with better insulation and reflective materials to reduce cooling demand. This is especially useful as temperatures rise and extreme heat events become more frequent.
    Early Warning Systems: Establish and upgrade storm warning systems to minimize the damage caused by severe weather, such as high winds and flash flooding. Integrating digital alert systems for residents will enhance preparedness for extreme weather.

2. Water Management

Roßtal, like many European towns, may face both droughts and floods due to climate change. A balanced water management system is essential.
Key Measures:

    Flood Management: Build or enhance flood protection infrastructure along rivers and flood-prone areas. This includes maintaining levees, and embankments, and developing wetland areas as natural flood buffers.
    Drought Resilience: Given potential water shortages, invest in efficient irrigation technologies for agriculture and encourage water conservation practices among residents. Collect and store rainwater for periods of drought, and modernize water infrastructure to reduce leakages.
    Sustainable Drainage Systems (SuDS): Develop SuDS to manage surface water runoff through swales, basins, and retention ponds that slow water movement and filter pollutants. These also provide green spaces and enhance biodiversity.

3. Energy Efficiency and Transition

Reducing energy consumption and transitioning to sustainable energy sources will benefit both climate adaptation and mitigation.
Key Measures:

    Energy-Efficient Buildings: Retrofit public and private buildings with energy-saving technologies, including LED lighting, solar panels, and energy-efficient HVAC systems. This reduces dependency on fossil fuels and prepares the town for potential heat-related power demands.
    District Heating Systems: Explore renewable-based district heating systems that utilize local biomass, geothermal energy, or waste heat to meet heating demand. Integrating renewable energy into the local grid will enhance resilience against energy shocks and reduce greenhouse gas emissions.
    Promote Solar Energy: Rooftop solar installations should be encouraged through local incentives, especially on public buildings and residential homes.

4. Agriculture and Food Security

Agriculture in Roßtal may be affected by altered growing seasons, increased droughts, and changes in pest patterns. Sustainable agriculture practices will ensure food security and resilience.
Key Measures:

    Diversification of Crops: Encourage local farmers to diversify crops and adopt drought-resistant varieties. Crop rotation and agroforestry practices can also build soil health and resilience against extreme weather.
    Water-Efficient Irrigation: Introduce water-saving irrigation methods, such as drip irrigation or sensor-based systems, to maximize water use during dry spells.
    Local Food Production: Support urban farming initiatives and local food markets to reduce the reliance on external food supply chains, which can be vulnerable to climate impacts elsewhere.

5. Ecosystem and Biodiversity Protection

Climate change will exert pressure on local ecosystems and biodiversity. Proactive management of natural areas can enhance resilience.
Key Measures:

    Conservation of Natural Habitats: Protect local forests and green spaces to maintain biodiversity, control erosion, and act as carbon sinks. Reforestation efforts with native species should be prioritized, along with efforts to restore degraded ecosystems.
    Biodiversity Corridors: Establish green corridors that allow wildlife to migrate and adapt to shifting habitats caused by climate changes.
    Water Body Restoration: Restore and protect local rivers, streams, and wetlands to maintain ecosystem health, provide flood mitigation, and ensure clean water supplies.

6. Community Engagement and Education

Effective adaptation requires local participation and an informed public.
Key Measures:

    Public Awareness Campaigns: Conduct climate literacy campaigns to educate residents about the risks of climate change and how they can contribute to adaptation efforts.
    Community-Based Adaptation Plans: Engage local communities in adaptation planning, emphasizing participatory approaches to identify vulnerabilities and prioritize local needs.
    Citizen Science Programs: Encourage community involvement in monitoring climate impacts, such as rainfall patterns, heatwave intensity, or biodiversity changes, which will provide valuable data for local climate adaptation strategies.

7. Policy and Governance

The adaptation strategy should be backed by strong governance frameworks and financial mechanisms to ensure long-term success.
Key Measures:

    Local Climate Action Plans: Develop and implement a municipal climate adaptation plan that sets specific targets, milestones, and funding mechanisms for adaptation projects.
    Public-Private Partnerships: Engage businesses in adaptation strategies, especially in sectors like energy, agriculture, and construction, where innovations can reduce the town’s vulnerability to climate risks.
    Funding for Climate Adaptation: Access European Union (EU) funds and national grants dedicated to climate adaptation, and explore local taxation or incentive programs to fund resilience projects.

Conclusion

Roßtal’s climate adaptation strategy must focus on mitigating immediate risks such as heatwaves and floods, while also enhancing long-term resilience through infrastructure upgrades, ecosystem protection, and community involvement. This integrated approach will help ensure that Roßtal can thrive in the face of climate change challenges.

\subsubsection{EcoScapes}

**Current State of Climate Adaptation in Rosstal and Surroundings: A Report**

**Introduction**

The town of Rosstal, nestled in a picturesque setting characterized by a mix of green fields, forests, and a well-developed public transportation system, presents a unique opportunity to explore climate adaptation strategies. This report aims to analyze the current state of climate adaptation in Rosstal, drawing insights from the RGB satellite image and moisture anomalies observed on a sunny day.

**Urban Layout and Green Spaces**

The image reveals Rosstal as a relatively small town, with a population ranging from 10,000 to 50,000 people. The town's layout is irregular, featuring a blend of buildings and green spaces. The central area is surrounded by expansive green fields, creating a sense of openness and tranquility. The presence of several forests and parks scattered throughout the city adds to its aesthetic and livability.

The city's size and the surrounding green fields highlight a balance between urban development and nature, offering residents and visitors a blend of urban convenience and natural beauty. The distribution of green spaces, including parks and gardens, serves as a vital component of climate adaptation, providing areas for cooling, recreation, and biodiversity conservation.

**Transportation Infrastructure**

The image depicts a well-developed transportation network, including railway lines and stations located in the middle of the city, connecting different parts of the urban area. The presence of two major bridges further enhances connectivity, facilitating movement between different regions. This infrastructure supports sustainable commuting options, reducing traffic congestion and promoting energy efficiency.

The transportation system's design and maintenance are crucial for climate adaptation, as they impact energy consumption and emissions. Efficient public transportation systems reduce the reliance on private vehicles, thereby lowering greenhouse gas emissions and contributing to a more sustainable urban environment.

**Moisture Anomalies and Heat Distribution**

The moisture anomalies map, though not provided in the reference, is a vital tool for understanding the microclimate of Rosstal. On a sunny day, the presence of blue and orange coloration in the image represents different moisture levels, indicating areas with higher or lower humidity.

The heat map, which would typically show areas with higher or lower temperatures, is crucial for identifying hotspots and cool zones within the town. These insights are essential for urban planning and infrastructure development, ensuring that buildings and public spaces are designed to mitigate the urban heat island effect.

**Conclusion**

Rosstal's current state of climate adaptation is characterized by a blend of natural and urban elements, a well-developed transportation system, and a mix of green spaces. The town's layout, with its central green fields and scattered forests, provides a foundation for effective climate adaptation strategies. The transportation infrastructure, while well-maintained, offers opportunities for further energy efficiency improvements.

The moisture anomalies and heat distribution, though not directly visualized in the provided image, are critical for understanding local climate conditions. These insights are necessary for designing buildings and public spaces that can withstand varying weather patterns, ensuring resilience and comfort for residents.

In summary, Rosstal presents a strong foundation for climate adaptation, with a mix of natural and urban elements, a well-developed transportation system, and a blend of green spaces. Continued efforts in urban planning, infrastructure development, and energy efficiency will be key to enhancing the town's resilience and sustainability in the face of climate change.

\subsubsection{EcoScapes + ChatClimate}

Climate Adaptation Strategy for Rosstal

Given Rosstal’s geographical context and the information provided on its urban layout, transportation infrastructure, and climate factors, this strategy focuses on three key climate adaptation areas: urban cooling, energy efficiency, and infrastructure resilience. Additionally, these recommendations are aligned with the latest scientific understanding of climate adaptation from the IPCC report.

1. Urban Cooling and Greening

Rosstal’s blend of green spaces and urban areas offers an excellent foundation for adaptation through nature-based solutions. These strategies would not only mitigate the urban heat island effect but also enhance biodiversity and improve residents' quality of life. Key recommendations include:

    Urban Tree Canopy Expansion: Increasing tree cover, particularly in urban areas and around transportation hubs, can help lower surface and air temperatures. A goal of increasing tree canopy coverage by 20\% would significantly reduce the heat island effect while improving air quality.

    Green Roofs and Walls: Implementing green roofs on public buildings and new developments can act as insulation, reducing energy costs. This strategy also provides additional green spaces, reduces runoff during heavy rains, and contributes to biodiversity by supporting pollinators and other species.

    Preservation of Existing Green Fields and Parks: Maintaining and enhancing the green fields surrounding the city can serve as cooling areas and create microclimates that buffer against temperature extremes. Ensuring that these spaces remain undeveloped is critical for long-term sustainability.

2. Energy Efficiency and Sustainable Urban Development

Rosstal’s existing public transportation system is a strength that should be leveraged for further climate resilience. Integrating energy-efficient technologies into the city’s infrastructure can drastically reduce carbon emissions and adapt to changing energy demands:

    Energy Efficiency in Buildings: Retrofitting existing buildings with energy-efficient technologies such as LED lighting, smart meters, and better insulation would reduce energy consumption and increase comfort for residents. New constructions should adhere to the highest energy efficiency standards (e.g., Passive House or nearly zero-energy buildings).

    Sustainable Transportation Infrastructure: Continue enhancing the public transportation network by adding more electric buses or light rail systems. Expanding charging infrastructure for electric vehicles would promote a shift from fossil-fuel-based transport to greener alternatives. This is essential given the well-developed rail network and bridges that support inter-regional travel.

    Decentralized Energy Systems: Encourage the adoption of rooftop solar panels, small-scale wind turbines, and community solar programs. Decentralized energy systems provide resilience during extreme weather events that could disrupt centralized grids.

3. Water Management and Infrastructure Resilience

With climate change predicted to increase the frequency of extreme weather events, including both droughts and heavy rainfall, Rosstal needs to ensure its infrastructure can withstand such variability.

    Water Retention and Stormwater Management: Implementing rainwater harvesting systems and permeable pavements in urban areas can help capture and store water during heavy rains. Green infrastructure solutions, such as bioswales and constructed wetlands, can further help manage stormwater while also providing habitats for wildlife.

    Moisture Anomalies Monitoring and Cooling Solutions: Integrating moisture sensors throughout the city will allow for real-time monitoring of dry or water-saturated areas. This data can help guide urban cooling strategies and irrigation for green spaces. In areas identified as “hotspots” through heat maps, interventions such as increased shading (e.g., with awnings or tree planting) and the installation of water features (fountains or misting systems) can help reduce temperatures.

4. Climate-Resilient Infrastructure

Given Rosstal’s well-developed transportation network, including major bridges and railway lines, resilience should be a priority to avoid disruptions during extreme weather:

    Climate-Proofing Critical Infrastructure: Railways and bridges should undergo climate risk assessments and retrofitting to ensure their resilience against flooding, extreme heat, or cold. Materials that can expand or contract with temperature variations should be prioritized for repairs.

    Heat-Resistant Pavement: Utilizing high-albedo materials in roads and public spaces can reduce surface temperatures and prolong the lifespan of infrastructure, reducing maintenance costs in the long term.

5. Community Engagement and Policy Integration

Successful climate adaptation requires active community engagement and a strong policy framework:

    Public Awareness Campaigns: Engage residents in the importance of climate adaptation through educational campaigns. This could cover energy conservation, tree planting programs, and the benefits of public transportation.

    Climate-Resilient Urban Planning: Integrating climate risk assessments into the town’s planning processes will ensure that future development is sustainable and adaptable. All new infrastructure projects should undergo a climate vulnerability assessment to identify potential risks and adaptation measures.

6. Policy and Governance

For the long-term success of these strategies, it’s critical to integrate climate adaptation into local governance:

    Local Climate Action Plan: Develop a climate action plan for Rosstal that sets out specific goals, targets, and timelines for implementing these adaptation measures. Aligning this plan with broader regional or national strategies will ensure funding opportunities and knowledge-sharing.

    Collaboration with Regional and National Authorities: Coordinate with higher government levels to access climate adaptation funding, technical support, and knowledge from larger-scale projects.

By focusing on urban cooling, energy efficiency, resilient infrastructure, and water management, Rosstal can become a model town for climate adaptation. These strategies will not only protect its infrastructure and residents from the impacts of climate change but also enhance the town’s sustainability and quality of life for future generations.

\subsection{Run 2 - Erlangen}

\subsubsection{Unchanged ChatClimate}

To develop a detailed climate adaptation strategy for Erlangen, we must consider several key aspects of its geographical, environmental, and socioeconomic context. Erlangen is located in Bavaria, Germany, and has a temperate climate with occasional flooding risks, and is subject to urban heat island (UHI) effects. Therefore, climate adaptation strategies for Erlangen should focus on addressing the city's specific vulnerabilities, especially concerning temperature rise, flood risks, and sustainable development goals.

1. Green Urban Infrastructure and Nature-based Solutions

    Urban Greening: Expanding green spaces can reduce the urban heat island effect, help manage stormwater, and improve biodiversity. Introducing green roofs, urban parks, and vertical gardens can reduce local temperatures while absorbing CO2. Prioritize planting drought-resistant and native species that can withstand projected climate changes.
    Wetland and River Restoration: Enhance natural flood buffers by restoring wetlands, managing the Regnitz River floodplain, and promoting riparian buffer zones. These measures will improve water quality, enhance biodiversity, and reduce flood risks.
    Tree Canopy Expansion: Expanding urban tree cover can help reduce ambient air temperatures and promote shade, reducing heat stress during summer heatwaves.

2. Water Management and Flood Control

    Sustainable Drainage Systems (SuDS): SuDS such as permeable pavements, green roofs, and rain gardens can help manage stormwater, reducing runoff and mitigating flood risks. These systems should be integrated into new developments and retrofitted where possible to control urban flash floods.
    Rainwater Harvesting: Encourage rainwater harvesting systems in both public and private buildings to reduce the reliance on mains water, particularly for non-potable uses like irrigation.
    Greywater Recycling: Implement greywater systems to reduce water waste, particularly in commercial and institutional settings, contributing to water conservation in a warming climate.

3. Energy Efficiency and Renewable Energy Transition

    Building Insulation and Retrofitting: Given projected increases in both heat and cold extremes, improving the energy efficiency of buildings through better insulation and retrofitting will be crucial. Passive cooling techniques such as natural ventilation, green roofs, and energy-efficient building materials will reduce the reliance on air conditioning and heating.
    District Heating Systems: Expand or enhance district heating networks powered by renewable energy sources such as geothermal or biomass, which are suitable for a region like Bavaria.
    Solar Energy Expansion: Promote rooftop solar PV systems for residential, commercial, and public buildings, reducing dependence on fossil fuels and lowering emissions.

4. Transportation and Mobility Solutions

    Sustainable Public Transport: Invest in expanding and improving the city's public transportation network, integrating electric buses, and promoting bicycle lanes. Providing incentives for electric vehicle (EV) adoption, including developing charging infrastructure, will also contribute to lowering transportation emissions.
    Pedestrian and Bicycle Infrastructure: Expand walkable and cycle-friendly infrastructure to reduce car dependency. A focus on active transport helps mitigate emissions and reduces urban heat through less road congestion.

5. Heat Stress and Public Health Adaptation

    Cooling Centers: Establish designated cooling centers across the city to provide refuge during extreme heat events, especially for vulnerable populations such as the elderly.
    Early Warning Systems: Implement heatwave early warning systems and create public awareness campaigns focused on the health risks associated with extreme heat and how to mitigate them.
    Green Health Strategies: Develop city health programs focusing on increasing resilience against vector-borne diseases and other health challenges exacerbated by climate change, such as respiratory diseases caused by poor air quality.

6. Policy, Governance, and Finance

    Climate-Resilient Zoning Laws: Update urban planning and zoning regulations to enforce climate-resilient building codes, prioritizing low-carbon materials and adaptive designs in new construction projects.
    Climate Bonds and Funding: Use climate bonds or leverage European Union green funds to finance sustainable infrastructure projects, such as flood defenses and renewable energy installations.
    Collaboration and Public Engagement: Engage local communities in adaptation planning by setting up citizen advisory panels. Collaboration with local universities and research institutions can also help integrate the latest climate science into urban development plans.

7. Biodiversity and Ecosystem Resilience

    Biodiversity Corridors: Create green corridors and connect urban green spaces to maintain biodiversity and allow species to migrate as the climate changes.
    Pollinator Habitats: Support biodiversity by protecting pollinator species through urban beekeeping projects, butterfly gardens, and pesticide-free zones.

By implementing these multi-sectoral strategies, Erlangen can build resilience to climate impacts, reduce carbon emissions, and create a healthier, more sustainable urban environment for its residents.

\subsubsection{EcoScapes}

**Current State of Climate Adaptation in Erlangen**

**Introduction**

Erlangen, a medium to large-sized city in Germany, has experienced significant urban development and infrastructure expansion. This report examines the current state of climate adaptation in Erlangen, focusing on the city's layout, urban infrastructure, green spaces, and thermal patterns, as derived from RGB satellite images and moisture anomaly descriptions.

**Urban Layout and Infrastructure**

The aerial view of Erlangen reveals a grid-like layout with a central city center surrounded by various buildings and infrastructure. The city's size and population are estimated to be in the medium to large range, characterized by a mix of residential, commercial, and industrial zones. The presence of multiple cars and trucks on the roads indicates a well-developed and accessible urban area.

Key features include:
- A large city center with a mix of residential, commercial, and industrial buildings.
- Well-developed roads and highways connecting different parts of the city.
- Multiple bridges and tunnels facilitating smooth movement of people and goods.
- Forests and parks surrounding the city, adding to its ecological balance.

**Green Spaces and Urban Planning**

Erlangen's green spaces, including forests and parks, play a crucial role in urban planning. These areas provide natural environments, enhance the city's aesthetic appeal, and contribute to ecological sustainability. The city's well-organized layout, with clear distinctions between urban and rural areas, supports effective climate adaptation strategies.

Key points:
- Forests and parks surrounding the city act as natural buffers.
- Green spaces within the urban area contribute to the overall well-being of residents.
- The city's layout supports efficient transportation and movement, reducing traffic congestion.

**Thermal Patterns and Urban Heat Island Effect**

The moisture anomaly map and thermal pattern analysis indicate a significant urban heat island effect in Erlangen. The presence of multiple red high heat spots, primarily in urban areas, suggests higher temperatures due to industrial zones, urban infrastructure, and lack of vegetation. These areas require targeted cooling systems and heat management strategies to mitigate the urban heat island effect.

Key observations:
- High heat levels in urban areas, likely due to industrial zones and urban infrastructure.
- Low heat levels in green spaces and rural areas, indicating potential cooling benefits.
- The need for heat island mitigation measures, including green spaces and reflective surfaces.

**Conclusion**

Erlangen's current state of climate adaptation reflects a mix of well-developed urban infrastructure and significant opportunities for further enhancement. The city's grid-like layout, extensive green spaces, and thermal patterns highlight both strengths and areas for improvement. By implementing targeted cooling systems, promoting green spaces, and addressing the urban heat island effect, Erlangen can enhance its climate resilience and provide a more comfortable urban environment for its residents.

\subsubsection{EcoScapes + ChatClimate}

Comprehensive Climate Adaptation Strategy for Erlangen
Introduction

Erlangen, a city with a mix of residential, commercial, and industrial zones, is experiencing the impacts of urban heat islands, infrastructure expansion, and significant urban development. The city is well-organized, surrounded by green spaces and parks, but urban areas show high heat anomalies, making climate adaptation a pressing need. This strategy leverages local geographic features and existing infrastructure to enhance climate resilience, reduce emissions, and improve the urban environment.

Key Challenges

    Urban Heat Island Effect: As highlighted by satellite data, urban areas in Erlangen exhibit higher temperatures, particularly around industrial zones and densely built neighborhoods.
    Moisture Anomalies: Discrepancies in moisture retention in urban versus rural areas, with urban regions suffering from heat stress due to a lack of vegetation and surface permeability.
    Traffic and Infrastructure: High road usage with cars and trucks contributing to emissions and heat, compounded by limited integration of green infrastructure within the city center.

Climate Adaptation Goals

    Reduce Urban Heat Island Effect: Implement measures to lower surface and air temperatures in high-heat zones.
    Enhance Green Spaces: Expand the city’s green spaces, especially in high-density urban areas, to increase cooling effects and ecological resilience.
    Sustainable Transportation: Reduce dependency on fossil fuels by promoting electric vehicles and enhancing public transportation systems.
    Water Management: Improve urban water retention through permeable surfaces and sustainable drainage systems.

Strategic Pillars

1. Urban Greening Initiatives

Urban greening will be central to Erlangen’s climate adaptation strategy, addressing both heat mitigation and ecological sustainability.

    Expand Urban Tree Canopy: Trees provide natural cooling through shading and evapotranspiration. Introducing tree-lined streets, especially in heat-prone industrial and residential zones, will help cool down urban areas.
    Vertical and Rooftop Gardens: Encourage rooftop and vertical gardens in residential and commercial buildings, particularly in the dense city center, to combat heat and improve air quality.
    Creation of Pocket Parks: Introduce small green spaces in heavily built-up areas to improve microclimates and offer respite from heat for local residents.

Potential Benefits: Reduced surface temperatures, better air quality, and improved mental well-being of residents.

2. Heat-Resistant and Reflective Urban Materials

Reducing the absorption of heat by urban surfaces is critical for mitigating the Urban Heat Island effect.

    Reflective Roofs and Pavements: Implement the use of reflective materials in roadways and building surfaces. Light-colored, heat-reflective materials can significantly reduce the heat absorption of structures and roadways.
    Cool Roofs for Industrial Zones: Promote the installation of "cool roofs" in industrial areas to reduce internal and external heat stress, lowering both emissions and energy demands for cooling systems.

Potential Benefits: Reduced indoor and outdoor temperatures, energy savings, and mitigation of heat-related health risks.

3. Sustainable Transportation

Given Erlangen’s well-developed road infrastructure, a transition toward greener transportation is necessary to reduce emissions and mitigate heat stress from traffic.

    Electric Vehicle (EV) Infrastructure: Expand the city’s EV charging infrastructure, incentivizing the use of electric vehicles. EV adoption is a key step in reducing transportation-related emissions.
    Enhance Public Transport and Cycling Networks: Promote public transportation and create a more connected cycling infrastructure. Integrated bus and cycling lanes will reduce car dependence, lower emissions, and improve urban air quality.
    Low Emission Zones: Implement low-emission zones in the city center to reduce traffic and encourage the use of cleaner transportation modes.

Potential Benefits: Improved air quality, reduced noise pollution, and better traffic management.

4. Water Management and Permeable Surfaces

Water retention and urban moisture management are crucial to addressing both heat stress and potential drought risks.

    Permeable Pavements: Introduce permeable materials in roadways and parking lots to allow water infiltration and reduce surface runoff. This will also help cool the city by reducing heat retention on paved surfaces.
    Rainwater Harvesting Systems: Encourage the installation of rainwater harvesting systems for both residential and commercial buildings. Captured rainwater can be used for irrigation, reducing the demand on the city’s freshwater supply during dry periods.
    Green Infrastructure for Drainage: Implement bioswales and rain gardens along roadways to capture runoff, filter pollutants, and enhance urban biodiversity.

Potential Benefits: Reduced flood risks, enhanced groundwater recharge, and cooler urban microclimates.

5. Energy Efficiency in Buildings

Erlangen’s building infrastructure must be upgraded to support lower energy consumption and improve climate resilience.

    Building Retrofitting for Energy Efficiency: Encourage retrofitting buildings with better insulation, energy-efficient windows, and smart cooling systems. These measures will reduce the need for air conditioning and heating, contributing to lower energy use and emissions.
    District Cooling Systems: Introduce district cooling systems, especially in the city’s commercial zones, to efficiently manage building temperatures and reduce individual air conditioning needs.

Potential Benefits: Lower energy costs, reduced emissions, and improved indoor air quality.

6. Community Engagement and Education

Engaging residents in climate adaptation is essential for long-term success.

    Public Awareness Campaigns: Initiate campaigns to educate residents about heat risks, water conservation, and the importance of green spaces. Encouraging local participation in tree planting and green roof programs can create a sense of shared responsibility for the city's climate resilience.
    Climate-Resilient Neighborhood Planning: Work with local communities to redesign neighborhoods with climate resilience in mind, incorporating green spaces, improved drainage systems, and sustainable mobility options.

Potential Benefits: Increased community resilience, stronger local participation in climate actions, and enhanced social cohesion.
Monitoring and Evaluation

    Heat Mapping: Continue to monitor thermal patterns in the city using satellite imagery to evaluate the effectiveness of heat mitigation strategies and adjust plans as necessary.
    Carbon Emissions Monitoring: Track the reduction in transportation and industrial emissions to measure progress in achieving climate adaptation goals.
    Water Usage Data: Regularly assess water consumption and runoff patterns to ensure the effectiveness of permeable surfaces and rainwater harvesting systems.

Conclusion

Erlangen’s climate adaptation strategy is designed to reduce the urban heat island effect, enhance the city’s green infrastructure, and promote sustainable energy and water usage. By focusing on integrated solutions, including green spaces, energy-efficient buildings, and sustainable transportation, the city can improve its resilience to climate change while creating a more livable environment for its residents.

\subsection{Run 3 - Roßtal}

\subsubsection{Unchanged ChatClimate}

Climate Adaptation Strategy for Rosstal

Introduction Rosstal is a small town in Bavaria, Germany, with a climate typical of Central Europe. As climate change intensifies, local communities such as Rosstal will need to adopt both mitigation and adaptation measures to cope with increasing environmental stresses. This strategy outlines a comprehensive approach based on the latest scientific research, tailored to the specific environmental and geographical characteristics of Rosstal.
1. Climate Profile and Projected Impacts

The IPCC AR6 Synthesis Report highlights key vulnerabilities in Europe due to climate change, including increased temperatures, irregular precipitation patterns, and a higher incidence of extreme weather events such as heatwaves and storms. Bavaria, including Rosstal, will likely experience:

    Warmer temperatures, with more frequent and severe heatwaves.
    Altered precipitation patterns leading to periodic droughts or intense rainfall, potentially overwhelming local water management systems.
    Increased strain on biodiversity and agriculture due to the changes in climate conditions.

2. Key Climate Adaptation Objectives

    Heat Mitigation: Address rising temperatures and heatwaves.
    Water Management: Secure a resilient water supply and drainage system to manage extreme rainfall and drought.
    Ecosystem Resilience: Enhance local biodiversity and ecosystem services.
    Energy Efficiency and Infrastructure Resilience: Improve energy systems and make infrastructure climate-resilient.

3. Specific Adaptation Measures
3.1 Urban Greening and Heat Management

    Urban Tree Canopy Expansion: Plant native and drought-resistant trees along streets and in public spaces to mitigate the urban heat island effect. Trees also act as carbon sinks and provide shading, reducing energy demand for cooling.
    Green Roofs and Walls: Promote green roofing systems on public buildings and new developments. These provide insulation, reduce runoff, and help cool the urban environment during heatwaves.
    Water Features and Reflective Surfaces: Install water bodies and fountains in public spaces to reduce ambient temperature. Implement the use of lighter-colored, reflective paving materials to reduce heat absorption.

3.2 Sustainable Water Management

    Rainwater Harvesting: Install systems to capture and store rainwater in public and private buildings. This would be particularly useful during dry periods, reducing demand on municipal water supplies.
    Green Infrastructure for Drainage: Introduce bioswales, rain gardens, and permeable pavements in urban areas to manage stormwater, reduce runoff, and limit the risk of flooding.
    Modernized Irrigation Systems: Equip agricultural areas around Rosstal with water-efficient irrigation systems such as drip irrigation to address periodic droughts. Promote on-farm water conservation techniques.

3.3 Ecosystem-Based Adaptation (EbA)

    Reforestation and Agroforestry: Integrate agroforestry practices on farms to improve soil health, water retention, and biodiversity. Planting trees along the edges of agricultural fields can also serve as windbreaks and reduce erosion.
    Wetland Restoration: Where possible, restore or create wetlands around Rosstal, which act as natural buffers against flooding, promote biodiversity, and enhance carbon sequestration.
    Biodiversity Corridors: Create and maintain green corridors to connect patches of forest and natural habitats, allowing wildlife to migrate in response to shifting climate conditions.

3.4 Energy Efficiency and Climate-Resilient Infrastructure

    Building Retrofits for Energy Efficiency: Encourage the retrofitting of public buildings and homes with better insulation, efficient heating systems, and energy-efficient windows. This will reduce energy consumption during both hot and cold periods.
    Renewable Energy: Expand the use of solar panels on residential and commercial buildings. Bavaria already has a strong renewable energy policy, and Rosstal can build on this by incentivizing rooftop solar panels, reducing dependency on fossil fuels.
    Resilient Transportation and Energy Infrastructure: Ensure that local transport networks (e.g., roads, rail, public transport) are resilient to extreme weather events, including floods and storms. Energy infrastructure, including power lines, should be upgraded to withstand these changes.

4. Governance, Engagement, and Education

    Community Involvement: Actively involve local residents and stakeholders in the planning and implementation of climate adaptation strategies through public consultations, educational workshops, and community greening projects.
    Climate Literacy and Awareness: Promote climate awareness through local schools and media campaigns, focusing on the importance of individual contributions to adaptation and mitigation efforts. Encourage sustainable behaviors such as energy conservation and waste reduction.

5. Monitoring and Evaluation

    Regular Climate Risk Assessments: Conduct regular assessments of climate risks to Rosstal, particularly regarding flood risks, heatwaves, and droughts. These assessments will help adjust the strategy as conditions evolve.
    Performance Indicators: Monitor progress using key indicators such as green space coverage, water use efficiency, renewable energy adoption rates, and reductions in heat-related health incidents.

6. Financing and Partnerships

    Leverage State and EU Funding: Take advantage of Bavaria’s state-level climate adaptation funds and European Union (EU) climate resilience programs to support large-scale projects like wetland restoration and urban greening.
    Public-Private Partnerships: Encourage partnerships between local government, private businesses, and non-profits to fund and implement projects such as green roofing or rainwater harvesting.

Conclusion

By adopting a forward-thinking, multi-faceted climate adaptation strategy, Rosstal can reduce its vulnerability to the impacts of climate change while improving the quality of life for its residents. This plan not only mitigates climate risks but also promotes sustainable development through green infrastructure, resilient ecosystems, and energy efficiency.

\subsubsection{EcoScapes}

\#\#\# Climate Adaptation Report for Rosstal

\#\#\#\# Executive Summary
Rosstal, a town with a population ranging from 10,000 to 50,000 people, presents a unique landscape characterized by a mix of urban and rural elements. This report analyzes the current state of climate adaptation in Rosstal, focusing on its urban layout, green spaces, transportation infrastructure, and heat distribution.

\#\#\#\# Urban Layout and Green Spaces
Rosstal's urban layout is irregular, featuring a blend of buildings and green spaces. The city center is situated in the middle, surrounded by large open fields, creating a sense of openness and tranquility. Several forests and parks are scattered throughout the city, contributing to its aesthetic and livability. These green spaces offer residents and visitors areas for relaxation and recreation.

\#\#\#\# Transportation Infrastructure
Rosstal boasts a well-developed public transportation system, including railway lines and stations located in the city center. Two major bridges connect different parts of the city, facilitating movement and reducing traffic congestion. This infrastructure supports sustainable commuting options, promoting environmental sustainability.

\#\#\#\# Heat Distribution
The RGB satellite image of Rosstal shows a blue and orange color scheme, indicating heat distribution across the city. Urban areas, represented by orange and yellow squares, have higher heat levels compared to the predominantly blue rural areas. This variation in heat levels is influenced by the density of buildings and infrastructure in urban areas, which trap heat and create microclimates. In contrast, rural areas with fewer structures and more open spaces experience lower heat levels.

\#\#\#\# Implications for Climate Adaptation
1. **Urban Planning and Green Spaces**: Rosstal's green spaces play a crucial role in mitigating the urban heat island effect. Enhancing these green areas through tree planting and urban parks can further improve thermal comfort and air quality.

2. **Transportation Infrastructure**: The existing transportation infrastructure supports sustainable commuting. Future investments should focus on enhancing public transportation options, reducing reliance on private vehicles, and promoting active transportation modes such as walking and cycling.

3. **Energy Consumption and Cooling Systems**: The higher heat levels in urban areas necessitate efficient energy consumption and cooling systems in buildings. Implementing energy-efficient building designs, such as passive cooling strategies and smart building technologies, can reduce energy costs and environmental impact.

4. **Resilience and Sustainability**: Rosstal's landscape, characterized by a mix of urban and rural elements, provides a natural buffer against extreme weather events. However, further resilience planning, including flood management and infrastructure upgrades, is essential to enhance the town's ability to withstand climate-related stressors.

\#\#\#\# Conclusion
Rosstal's current state of climate adaptation reflects a balanced approach to urban development and environmental stewardship. The town's mix of green spaces, efficient transportation infrastructure, and heat management strategies position it well for future climate challenges. Continued investments in sustainable urban planning, energy efficiency, and resilience will be key to ensuring Rosstal remains a comfortable and sustainable living environment in the face of climate change.

\subsubsection{EcoScapes + ChatClimate}

Detailed Climate Adaptation Strategy for Rosstal

Introduction Rosstal, characterized by a blend of urban and rural environments, faces distinct climate adaptation challenges and opportunities. With its small to medium-sized population (10,000-50,000), diverse green spaces, developed transportation infrastructure, and varying heat distribution, Rosstal's adaptation to climate change must enhance its resilience while capitalizing on its existing strengths.

1. Urban Greening and Landscape Management

Urban greening is central to mitigating the urban heat island (UHI) effect, which is significant in Rosstal due to the higher heat levels in built-up areas.

    Green Infrastructure Expansion: Urban parks and street trees should be expanded, especially in areas with high UHI. These spaces will cool surrounding areas, provide shade, and absorb carbon dioxide. Trees like oaks, elms, and lindens, which are drought-tolerant, can reduce local temperatures and improve air quality.

    Forest Preservation: Rosstal already benefits from scattered forests. To enhance climate resilience, forest areas should be protected from urban encroachment and sustainably managed to prevent soil erosion, sequester carbon, and maintain biodiversity.

    Green Roofs and Walls: Incentivizing green roofs and walls on public and private buildings will help absorb rainwater, provide insulation, and cool the surrounding environment.

Co-benefits: Urban greening will improve mental health and provide spaces for social engagement and physical activity. It will also improve biodiversity and reduce surface runoff during heavy rainfalls.

2. Heat Mitigation through Sustainable Urban Design

Given Rosstal’s higher heat levels in urban areas:

    Passive Cooling Design: New and retrofitted buildings should prioritize passive cooling techniques, such as optimized window shading, reflective materials, and natural ventilation, to reduce energy consumption for air conditioning.

    Cool Pavements and Surfaces: Introduce reflective or porous materials in streets and parking lots to reduce heat absorption. This will lower surface temperatures and reduce heat transfer to the surrounding environment.

    Shaded Public Spaces: Develop shaded outdoor areas, particularly near transportation hubs and city centers, to offer relief during heat waves and provide gathering spaces.

Co-benefits: Reducing dependency on air conditioning will cut energy use and emissions, contributing to both climate mitigation and adaptation goals.

3. Resilient Transportation and Mobility

Rosstal has a strong transportation infrastructure, but future climate stressors will require enhanced resilience.

    Sustainable Public Transport: Electrification of buses and trains will reduce emissions and limit the urban heat generated by fossil fuel engines. Moreover, establishing shaded or green corridors around railways and major roads can reduce heat absorption.

    Promoting Active Transportation: To reduce reliance on cars, create more pedestrian and cycling infrastructure. Green pathways can connect residential areas with city centers and parks, reducing emissions and promoting public health.

    Flood-Resilient Infrastructure: Since transportation networks are often vulnerable to extreme weather, invest in elevating rail and bridge infrastructure, and installing permeable materials to ensure resilience against potential flash floods.

Co-benefits: Sustainable transportation improves air quality, reduces traffic congestion, and promotes healthier lifestyles. Furthermore, energy-efficient infrastructure investments will boost economic resilience.

4. Energy Efficiency and Building Adaptation

Efficient energy use is critical, especially considering the heightened heat levels in urban zones.

    Building Energy Efficiency: Promote energy-efficient retrofitting of buildings, such as insulation, double-glazed windows, and renewable energy installations (e.g., rooftop solar panels). Energy efficiency reduces the need for cooling and heating, contributing to lower emissions and energy costs.

    Smart Building Technologies: Invest in smart grids and building management systems that optimize energy use based on real-time demand. These systems can reduce the burden on the electrical grid during heat waves and contribute to overall energy resilience.

Co-benefits: Implementing these measures will reduce energy costs for households and businesses, improve indoor comfort, and support local job creation in retrofitting and renewable energy sectors.

5. Water Resource Management and Resilience

Although Rosstal's climate adaptation report does not mention current flooding risks, resilient water management is crucial in the face of potential changes in precipitation patterns.

    Rainwater Harvesting: Encourage the installation of rainwater harvesting systems on public and private properties. This can reduce pressure on local water supplies during dry periods and mitigate runoff during heavy rains.

    Permeable Surfaces: Increase the use of permeable materials in urban development to enhance groundwater recharge and prevent stormwater runoff, which can cause flooding and overwhelm drainage systems.

    Drought-Resilient Agriculture: Farmers in the surrounding rural areas can adopt agroecological practices, including crop diversification and water-efficient irrigation systems, to adapt to changing rainfall patterns and reduce the vulnerability of food systems.

Co-benefits: Sustainable water management will protect the town’s freshwater resources, support agricultural productivity, and reduce risks of water scarcity.

6. Community Engagement and Climate Literacy

For any adaptation strategy to be effective, community involvement and awareness are essential.

    Climate Education Programs: Launch local campaigns to raise awareness about climate risks, the importance of green spaces, energy savings, and sustainable transportation. Schools and community centers can offer workshops to promote climate literacy and empower residents to take individual action.

    Citizen Participation: Establish citizen advisory committees to involve the public in decision-making processes related to climate adaptation. Engaging residents in urban greening projects or energy-saving initiatives can build a stronger community resilience network.

Co-benefits: A well-informed and engaged population is more likely to adopt sustainable practices, contributing to the long-term success of climate adaptation efforts.

7. Disaster Preparedness and Early Warning Systems

Given that extreme weather events (e.g., heat waves, heavy precipitation) are expected to become more frequent:

    Early Warning Systems: Invest in localized, real-time weather monitoring and early warning systems for heat waves, storms, and potential droughts. The system should be easily accessible via public platforms and mobile apps.

    Emergency Response Plans: Strengthen local emergency services’ capacity to respond to climate-related disasters. This should include heat shelters, cooling centers, and safe evacuation routes during extreme weather events.

Co-benefits: Effective disaster preparedness reduces human suffering, minimizes economic losses, and strengthens overall resilience to climate shocks.

8. Conclusion

Rosstal’s climate adaptation strategy should capitalize on its mix of urban and rural landscapes, leveraging natural assets like forests and green spaces to reduce heat, enhance resilience, and promote sustainability. Integrated investments in green infrastructure, energy efficiency, sustainable transport, and community engagement will create a robust, resilient environment for Rosstal’s future while contributing to climate mitigation efforts.

For further details on the scientific foundations supporting these measures, refer to the IPCC AR6 synthesis report, which highlights the urgency of rapid and coordinated adaptation to avoid escalating impacts from climate change.

\subsection{Run 3 - Erlangen}

\subsubsection{Unchanged ChatClimate}

Climate Adaptation Strategy for Erlangen, Germany

Erlangen, like many urban areas in Germany, faces a variety of climate challenges, including rising temperatures, increasing frequency of heatwaves, and changes in precipitation patterns, which can lead to droughts and intensified flooding events. To adapt to these challenges, the city can implement a comprehensive and scientifically informed strategy based on the latest insights from the IPCC AR6 Synthesis Report (2023) and regional climate studies.

1. Urban Greening and Nature-Based Solutions

Urban areas are particularly vulnerable to heatwaves and reduced air quality, which are expected to worsen with climate change. The IPCC highlights the importance of urban greening to mitigate heat and provide co-benefits for air quality, health, and biodiversity.

    Increase Green Spaces: Expand parks, community gardens, and green corridors to reduce the urban heat island effect and improve local air quality.
    Rooftop and Vertical Gardens: Encourage green roofs and walls to insulate buildings, reduce energy consumption, and sequester carbon.
    Street Trees and Green Canopies: Plant more street trees in highly populated areas to provide shade, reduce surface temperatures, and promote biodiversity.
    Wetland Restoration: Enhance natural flood control systems by restoring wetlands along rivers and low-lying areas prone to flooding.
    
2. Water Management and Resilience

Erlangen is expected to face both increased droughts and flooding risks due to changing precipitation patterns. A resilient water management strategy is crucial to cope with both extremes.

    Rainwater Harvesting and Recycling: Encourage the installation of rainwater harvesting systems in residential and commercial buildings for non-potable uses such as irrigation and sanitation.
    Permeable Pavements: Use permeable materials in sidewalks and parking lots to enhance groundwater recharge and reduce surface runoff during heavy rains.
    Floodplain Management: Implement zoning regulations that restrict construction in flood-prone areas, and enhance the capacity of the Pegnitz River to handle increased floodwaters by maintaining natural buffers.
    Smart Water Infrastructure: Invest in modern infrastructure that can store excess water during heavy rainfall events and redistribute it during droughts.

3. Building Energy Efficiency and Cooling

Buildings in Erlangen are expected to experience higher cooling demands as temperatures rise. Retrofitting buildings for energy efficiency will be key in reducing energy use and costs, while maintaining comfort for residents.

    Energy-Efficient Retrofits: Upgrade existing buildings with better insulation, double-glazed windows, and reflective roofing materials to reduce the need for heating and cooling.
    Passive Cooling Systems: Encourage the use of shading devices, ventilation systems, and green roofs to reduce indoor temperatures without energy-intensive air conditioning.
    District Cooling Systems: Explore district cooling options where excess heat from industries or renewable sources can be harnessed to cool nearby homes and buildings.

4. Sustainable Mobility and Transportation

Transportation is a key sector that can contribute to both climate mitigation and adaptation in Erlangen. The city should focus on promoting low-emission and climate-resilient transport systems.

    Cycling and Pedestrian Infrastructure: Invest in safe, extensive cycling lanes and pedestrian pathways to promote active transport, which reduces emissions and improves public health.
    Electric Mobility: Expand electric vehicle (EV) charging infrastructure and incentivize the use of EVs, buses, and shared mobility services.
    Public Transportation Improvements: Ensure the public transport system is energy-efficient, climate-resilient, and accessible. Invest in electric buses and improve rail connections to reduce reliance on personal vehicles.

5. Energy Transition and Renewable Sources

The transition to renewable energy sources is crucial in both mitigating climate impacts and building resilience in the face of changing energy demands.

    Solar Power Expansion: Encourage the installation of solar photovoltaic (PV) systems on rooftops and in open spaces to meet local energy demands sustainably.
    District Heating with Renewable Sources: Expand district heating systems that utilize biomass, geothermal, or solar energy to reduce reliance on fossil fuels.
    Energy Storage Systems: Invest in battery storage and decentralized grids to ensure energy security during peak demand or supply disruptions caused by extreme weather events.

6. Climate-Smart Agriculture and Urban Food Systems

Although Erlangen is an urban area, its proximity to agricultural lands means it is important to ensure that local food production is resilient to climate changes.

    Support Sustainable Agriculture: Promote climate-resilient agricultural practices in nearby rural areas, such as water-efficient irrigation, drought-tolerant crops, and agroforestry systems.
    Urban Agriculture: Encourage urban farming initiatives, community gardens, and vertical farming in unused urban spaces to strengthen local food systems.
    Food Waste Reduction: Implement programs to reduce food waste across the supply chain, from production to consumption, and promote composting and recycling of organic waste.

7. Community-Based Adaptation and Education

Engaging the local community in climate adaptation efforts is essential for ensuring long-term success.

    Public Awareness Campaigns: Launch educational initiatives to raise awareness of the importance of water conservation, energy efficiency, and climate-friendly transportation.
    Community Resilience Plans: Work with local neighborhoods to develop community-specific climate adaptation plans, focusing on vulnerable populations such as the elderly or low-income groups.
    Climate Adaptation Workshops: Offer training and workshops on sustainable practices such as rainwater harvesting, urban gardening, and energy-saving techniques.

8. Governance and Policy Frameworks

To implement these adaptation strategies, strong governance and policy support are essential.

    Integrate Climate Considerations into Urban Planning: Ensure that all new development projects consider long-term climate impacts, including flood risks, heat stress, and energy needs.
    Climate Risk Assessments: Conduct regular climate vulnerability assessments and integrate the findings into decision-making processes for infrastructure investments.
    Funding for Climate Initiatives: Secure funding from local, national, and EU-level programs to support adaptation initiatives. Partner with private sector stakeholders to co-finance projects that align with climate resilience goals.

Conclusion

By implementing this multi-faceted climate adaptation strategy, Erlangen can become a more resilient and sustainable city. These actions will not only reduce the risks posed by climate change but also improve the quality of life for its residents by creating greener, healthier, and more sustainable urban environments.

\subsubsection{EcoScapes}

\#\#\# Report on the Current State of Climate Adaptation in Erlangen

\#\#\#\# Introduction
Erlangen, a medium to large-sized city, showcases a blend of urban development and natural landscapes, with a grid-like layout and diverse infrastructure. The city's climate adaptation strategies are multifaceted, addressing both urban heat island effects and the integration of green spaces to enhance ecological balance and resilience.

\#\#\#\# Urban Layout and Infrastructure
The city's aerial view reveals a well-developed urban landscape with a mix of residential, commercial, and industrial areas. The central hub is surrounded by various buildings and infrastructure, including roads and highways, indicating a significant population and economic activity. The grid-like arrangement of streets and buildings facilitates efficient transportation and movement, enhancing connectivity and functionality.

\#\#\#\# Green Spaces and Forests
Erlangen is enveloped by a large forest, which acts as a natural buffer zone, enhancing the city's ecological balance. The presence of parks and green spaces within the urban area contributes to the overall aesthetics and well-being of residents. These green spaces help mitigate the urban heat island effect, providing cooler microclimates within the city.

\#\#\#\# Public Transportation
The city boasts a well-developed public transportation system, with railway lines running through the urban area. This network of railway lines and stations ensures efficient transportation for residents and visitors, reducing reliance on personal vehicles and contributing to lower greenhouse gas emissions.

\#\#\#\# Bridges and Tunnels
The city's infrastructure includes several major bridges and tunnels, enhancing connectivity and facilitating smooth movement of people and goods. These structures are crucial for managing traffic flow and ensuring accessibility across different parts of the city.

\#\#\#\# Urban Heat Island Effect
The thermal map of Erlangen highlights areas with varying heat levels, indicating the presence of urban heat island effects. High heat levels are often associated with industrial zones, buildings with dark roofs, and areas with less vegetation. To mitigate these effects, the city can focus on increasing green spaces, using reflective materials, and implementing urban cooling strategies.

\#\#\#\# Water Bodies
The presence of large expanses of greenery or parkland within the city suggests the integration of water bodies, such as lakes or rivers, which naturally provide cooler temperatures compared to land surfaces. These water bodies contribute to the overall ecological balance and provide recreational areas for residents.

\#\#\#\# Conclusion
Erlangen's climate adaptation strategies are well-aligned with its urban development and natural landscapes. The city's focus on green spaces, efficient transportation, and mitigation of the urban heat island effect positions it as a model for sustainable urban planning. Continued investment in green infrastructure, smart technologies, and community engagement will further enhance the city's resilience and livability in the face of climate change.

\subsubsection{EcoScapes + ChatClimate}

Comprehensive Climate Adaptation Strategy for Erlangen

1. Introduction

Erlangen, a medium-sized city with a mix of urban development and natural areas, is addressing the challenges posed by climate change through its existing climate adaptation strategies. With a focus on managing urban heat, increasing green spaces, and reducing carbon emissions, the city is well-positioned to enhance its resilience. However, further steps are needed to align with the latest global best practices in climate adaptation.

This strategy proposes a multi-faceted approach, incorporating measures across urban planning, infrastructure, and social systems to ensure the city's long-term resilience.

2. Urban Layout and Infrastructure

2.1 Urban Heat Island Mitigation

The city’s grid-like infrastructure contributes to heat retention, exacerbating the Urban Heat Island (UHI) effect. A targeted cooling plan is essential for improving livability during increasingly hot summers. Measures include:

    Green Roofs and Cool Roofing Materials: Encourage or mandate reflective surfaces and green roofs on both new and retrofitted buildings. Cool materials with high albedo reduce absorbed heat, mitigating the UHI effect. Incentives should be provided for industrial zones, where heat retention is higher.
    Shading and Cool Pavements: Replace dark asphalt with cool pavements that absorb less heat and maintain ambient temperatures. This can be combined with increased shading from trees or artificial canopies along pedestrian-heavy streets.

2.2 Sustainable Transportation Infrastructure

Erlangen’s efficient public transport system already contributes to reducing greenhouse gas emissions. To further enhance climate resilience:

    Expand Electric Public Transport: Gradually replace fossil fuel-powered buses and trains with electric vehicles (EVs) or hydrogen-powered models. Investment in charging infrastructure for electric buses and incentives for electric personal vehicles should complement this shift.
    Bicycle and Pedestrian Infrastructure: Enhance and expand bicycle lanes and pedestrian zones to further reduce vehicle emissions and improve air quality. Priority zones near schools, parks, and shopping areas should be targeted for new bicycle infrastructure.

3. Green Spaces and Ecosystem Services

3.1 Urban Green Spaces

Erlangen's existing parks and proximity to large forests are vital for ecological balance and climate adaptation. However, these areas need expansion and management to improve their climate resilience functions:

    Tree Planting and Green Corridors: Increase urban tree coverage by at least 10\% in residential and commercial areas. Green corridors connecting city parks to surrounding forests will facilitate biodiversity, improve air quality, and mitigate heat.
    Parks as Cooling Zones: Designate parks as ‘cooling zones,’ offering relief during heatwaves. Strategies include denser tree canopies and water features, which lower local temperatures and provide shade for residents.

3.2 Ecosystem-Based Adaptation

Forests surrounding Erlangen provide natural cooling and carbon sequestration. Integrating Ecosystem-based Adaptation (EbA) approaches can enhance these services:

    Forest Buffer Zones: Expand the city's forest buffer zones, prioritizing native species and enhancing the forest’s ability to absorb floodwaters during intense rainfall events.
    Wetland Restoration: If water bodies are present, restore degraded wetlands to act as natural flood mitigation systems and provide cooling through evapotranspiration.

4. Water Management and Flood Resilience

4.1 Green Infrastructure for Flood Control

With potential increases in extreme weather events due to climate change, the integration of sustainable urban drainage systems (SUDS) is crucial:

    Permeable Pavements and Bioswales: Install permeable pavements and bioswales throughout the city to capture and slowly release stormwater, reducing flood risks and enhancing groundwater recharge.
    Rain Gardens and Retention Ponds: Encourage the development of rain gardens and retention ponds in residential areas to capture runoff and prevent urban flooding.

4.2 Water Conservation

With global predictions of more intense droughts, efficient water management systems will be key:

    Greywater Recycling: Promote greywater recycling systems in both residential and commercial buildings to reduce fresh water consumption and increase water availability during dry periods.
    Smart Irrigation Systems: Implement smart irrigation in parks and public gardens to optimize water usage and prevent wastage.

5. Energy Efficiency and Carbon Neutrality

5.1 Energy-efficient Buildings

Increasing the energy efficiency of buildings is critical to reducing carbon emissions and enhancing climate resilience. Strategies include:

    Passive Design Techniques: Promote the use of passive solar design, natural ventilation, and better insulation in new buildings and renovations to reduce energy consumption for heating and cooling.
    Building Retrofit Programs: Implement a city-wide program to retrofit older buildings with energy-efficient windows, insulation, and solar panels. A combination of financial incentives and regulations could accelerate uptake.

5.2 Renewable Energy Integration

Erlangen can enhance its energy resilience by expanding its reliance on renewable energy sources:

    Solar and Wind Energy Expansion: Expand the city’s solar capacity by promoting rooftop solar installations on public, commercial, and residential buildings. Additionally, explore the feasibility of small-scale wind energy projects in less urbanized areas.
    District Heating Systems: Explore district heating systems powered by renewable energy sources to provide heating and cooling in dense urban areas, reducing reliance on fossil fuels.

6. Community Engagement and Governance

6.1 Community-based Adaptation

Effective climate adaptation requires active participation from local communities. Key actions include:

    Climate Education Campaigns: Launch public campaigns to raise awareness about climate change impacts and the importance of energy efficiency, water conservation, and sustainable transport. Partner with schools and local organizations to enhance outreach.
    Incentives for Sustainable Practices: Offer subsidies or tax breaks for homeowners who install energy-efficient appliances, green roofs, or solar panels, and businesses that implement sustainability measures.

6.2 Climate Resilient Governance

Strengthening governance mechanisms will be critical to successful climate adaptation:

    Climate Resilience Office: Establish a dedicated city office to coordinate climate adaptation efforts across sectors. This office should manage urban planning, public infrastructure, and community engagement to ensure all projects align with the city’s long-term adaptation goals.
    Climate Action Monitoring: Develop a monitoring and reporting system to track the city’s progress on climate adaptation and resilience, enabling data-driven adjustments to the strategy.

7. Conclusion

Erlangen has a solid foundation in climate adaptation, but greater efforts are required to meet future challenges. By integrating green infrastructure, enhancing energy efficiency, and engaging the community, Erlangen can strengthen its climate resilience, improve livability, and contribute to global efforts to mitigate climate change. This strategy provides a roadmap for continued leadership in sustainable urban development.

\subsection{Run 4 - Roßtal}

\subsubsection{Unchanged ChatClimate}

To develop a climate adaptation strategy for Roßtal, we can focus on several key areas that consider both its current climate vulnerabilities and long-term resilience. Roßtal, like many small towns in Central Europe, faces potential impacts from increased temperature extremes, shifts in precipitation patterns, and more frequent heatwaves, which can affect agriculture, infrastructure, and public health. Based on the latest guidance from climate research (including IPCC reports) and successful adaptation strategies implemented globally, here is a detailed strategy tailored for Roßtal:

1. Urban Greening and Ecosystem-based Adaptation

    Tree Canopy Expansion: Increase tree coverage throughout the town, particularly along streets, parks, and public spaces, to provide shade, reduce heat island effects, and improve air quality. Urban trees also contribute to reducing runoff during heavy rainfall events.
    Green Infrastructure: Encourage the development of green roofs and vertical gardens on public and private buildings. These can reduce building temperatures, decrease energy demand for cooling, and improve stormwater management.
    Restoration of Local Ecosystems: Identify and protect local ecosystems, such as riverbanks and forested areas. Restoring natural landscapes around Roßtal can enhance biodiversity and reduce flood risks by promoting natural water absorption and slowing down surface runoff.

2. Water Management and Resilience

    Sustainable Urban Drainage Systems (SUDS): Implement systems that manage heavy rainfall, such as permeable pavements, rain gardens, and retention ponds. These solutions can help manage increasing precipitation variability and avoid flash floods.
    Rainwater Harvesting: Encourage households and businesses to install rainwater harvesting systems for non-potable uses (e.g., irrigation, toilet flushing). This reduces reliance on water supplies during drought periods, which may become more frequent.
    Upgrading Water Infrastructure: Ensure that the town’s drainage systems and wastewater infrastructure are modernized to handle more intense rainfalls. This is critical for preventing localized flooding, especially in low-lying areas of Roßtal.

3. Energy Efficiency and Climate-Resilient Buildings

    Passive Cooling and Building Design: Promote the use of energy-efficient building designs that reduce the need for mechanical cooling. This includes the use of proper insulation, shading, reflective materials, and natural ventilation systems.
    Energy-Efficient Retrofitting: Encourage retrofitting of old buildings with energy-efficient technologies (e.g., insulation, high-performance windows, solar panels). Given that Roßtal may experience hotter summers, reducing the need for energy-intensive air conditioning is critical.
    Solar Energy Expansion: Incentivize the installation of solar panels on rooftops and in open spaces, especially in public buildings. Solar energy offers a renewable alternative that can mitigate the town's reliance on fossil fuels, contributing to both mitigation and adaptation goals.

4. Heatwave Preparedness and Public Health

    Public Cooling Centers: Establish cooling centers in public areas where vulnerable populations, such as the elderly, can access air-conditioned environments during extreme heat events.
    Early Warning Systems: Develop an early warning system for heatwaves, combined with public health campaigns to educate the community on heat-related risks, hydration, and heatstroke prevention.
    Urban Cooling Corridors: Create green corridors throughout the town to ensure airflow and reduce heat buildup. Such corridors should connect green spaces and be designed to optimize cooling during hot weather.

5. Agriculture and Food Security

    Climate-Resilient Crops: Promote the use of climate-resilient crop varieties that can withstand drought and heat stress, ensuring local agricultural productivity remains stable. Farmers in Roßtal should also be encouraged to adopt sustainable farming techniques like agroforestry and conservation tillage.
    Water-Efficient Farming: Invest in modern irrigation technologies such as drip irrigation, which optimizes water use in agricultural fields. This can prevent water wastage and improve resilience to periods of low rainfall.
    Local Food Networks: Support local food production through community-supported agriculture (CSA) programs and farmers' markets. This strengthens food security and reduces dependence on external food sources, which might be disrupted due to global climate shifts.

6. Disaster Risk Reduction

    Flood Risk Mapping and Zoning: Conduct comprehensive flood risk mapping for Roßtal, particularly for areas near rivers or low-lying regions. Zoning regulations should ensure that new developments are built outside of flood-prone areas.
    Flood Defenses: Invest in natural and engineered flood defenses, such as levees or buffer zones along rivers. Wetland restoration projects can also play a role in absorbing excess water during heavy rainfall events.
    Community Preparedness: Implement community-wide disaster preparedness programs that educate residents on how to respond to extreme weather events, such as floods and storms, and establish community response teams for emergency situations.

7. Climate Governance and Engagement

    Climate Adaptation Plan: Establish a local climate adaptation and mitigation committee to oversee the implementation of these strategies, working closely with local stakeholders, residents, and businesses.
    Climate Literacy and Public Participation: Promote educational campaigns that raise awareness about climate risks and solutions in Roßtal. Engage the public in decision-making through workshops, participatory planning processes, and local consultations.
    Sustainable Mobility: Encourage walking, cycling, and the use of public transportation to reduce greenhouse gas emissions and create a more resilient transportation network in the town.

8. Finance and Policy Support

    Funding Mechanisms: Secure funding for adaptation measures through both local budget allocations and external sources (e.g., European Union climate adaptation funds). Engage in partnerships with regional authorities and international organizations to attract investments.
    Regulations and Incentives: Implement local policies that require or incentivize sustainable building practices, water conservation, and renewable energy use. Offer tax benefits, subsidies, or grants to citizens and businesses adopting climate-smart technologies.

This integrated climate adaptation strategy for Roßtal aligns with the principles outlined in the IPCC AR6 report, emphasizing the need for urgent, multi-sectoral action to reduce risks, build resilience, and ensure a sustainable future.

\subsubsection{EcoScapes}

**Current State of Climate Adaptation in Rosstal**

**Introduction**

The town of Rosstal, situated in a picturesque setting surrounded by green fields and forests, presents a unique opportunity to explore the current state of climate adaptation. This report aims to analyze the town's adaptation strategies based on its urban layout, green spaces, transportation infrastructure, and heat distribution patterns, as derived from the RGB satellite image and moisture anomaly data.

**Urban Layout and Green Spaces**

Rosstal's layout is characterized by a mix of buildings and green spaces, with a central area surrounded by expansive green fields. The presence of several forests and parks scattered throughout the city adds to its aesthetic and livability. The green spaces provide essential natural buffers, contributing to the town's resilience against extreme weather events and heat waves.

**Transportation Infrastructure**

The town boasts a well-developed transportation system, including railway lines and stations in the center, facilitating easy access to the urban area. The presence of two major bridges further enhances connectivity, allowing for efficient movement of people and goods. This infrastructure supports sustainable commuting options, reducing traffic congestion and promoting eco-friendly transportation.

**Heat Distribution and Climate Adaptation**

The heat map, represented in a blue and orange color scheme, reveals a gradient of temperatures across Rosstal. The urban areas, characterized by orange and yellow squares, exhibit higher heat levels, while rural areas are predominantly blue, indicating cooler temperatures. This variation in heat levels can be attributed to the higher density of buildings and infrastructure in urban areas, which trap heat and create microclimates. In contrast, rural areas benefit from better air circulation and more open spaces, resulting in lower heat levels.

The presence of high heat spots, indicated by red areas, could be due to industrial activities, lack of vegetation, or natural phenomena. Conversely, low heat spots, represented by blue areas, are likely influenced by water bodies, dense vegetation, or areas with lower heat conductivity. Understanding these heat patterns is crucial for implementing effective climate adaptation strategies, such as incorporating heat-resistant materials in building construction and designing green spaces to mitigate the heat island effect.

**Conclusion**

Rosstal's current state of climate adaptation is characterized by a balanced integration of urban development and natural green spaces. The town's transportation infrastructure supports sustainable commuting, reducing traffic congestion and environmental impact. The analysis of heat distribution patterns provides valuable insights for further enhancing climate resilience through targeted urban planning and green space development. By addressing the unique challenges presented by its heat distribution, Rosstal can foster a resilient and sustainable urban environment for its residents.

**Recommendations**

1. **Enhance Green Spaces**: Invest in additional green spaces, particularly in urban areas, to provide natural cooling effects and improve air quality.
2. **Sustainable Building Practices**: Promote the use of heat-resistant materials and design buildings with natural ventilation to reduce energy consumption and improve thermal comfort.
3. **Public Awareness and Education**: Educate residents on climate adaptation measures, including energy conservation practices and the importance of maintaining green spaces.
4. **Monitoring and Data Analysis**: Continuously monitor heat distribution patterns and moisture anomalies to refine climate adaptation strategies and respond to emerging challenges.

By implementing these recommendations, Rosstal can strengthen its climate adaptation efforts, ensuring a resilient and livable urban environment for future generations.

\subsubsection{EcoScapes + ChatClimate}

To create a comprehensive climate adaptation strategy for Rosstal, it’s important to consider the specific characteristics of the town, as well as general climate adaptation measures derived from scientific findings. Based on the provided report on Rosstal and insights from the IPCC's Sixth Assessment Report, here is a detailed adaptation strategy for the town:

1. Enhancing Green Infrastructure and Natural Buffers

Rosstal benefits from its surrounding green fields, forests, and parks, which act as natural buffers against climate impacts such as heat waves and extreme weather events. However, increasing urbanization and changing climate patterns could reduce their effectiveness. Key actions include:

    Expand Urban Green Spaces: Given the town's heat distribution, with hotter areas in urban zones, increasing urban green spaces, particularly near areas with dense buildings, will help reduce the urban heat island (UHI) effect. Tree planting along streets, green roofs, and vertical gardens will enhance natural cooling.
    Forest and Wetland Conservation: Protecting and expanding surrounding forests, especially near waterways and urban edges, can absorb stormwater, reduce heat, and enhance air quality. Wetland restoration will also provide critical moisture regulation, particularly during droughts.

Justification: Studies have shown that ecosystem-based adaptation (EbA) approaches like forest restoration and urban greening can significantly reduce flood risks, urban heat, and increase biodiversity.

2. Sustainable Building Practices

With rising temperatures and increased likelihood of heatwaves, adapting Rosstal’s buildings for resilience is essential. This includes both new constructions and retrofits to existing structures:

    Use of Heat-Resistant Materials: Promote the use of light-colored, reflective materials for roofs and pavements to reduce heat absorption. Incorporate natural ventilation systems and shading structures to reduce energy demand for cooling.
    Green Building Standards: Implement or update building codes to require energy-efficient designs, including insulation, energy-saving appliances, and passive solar heating to reduce overall emissions and improve resilience during extreme heat events.

Justification: Building design plays a crucial role in energy consumption and heat mitigation. Adopting standards like these can reduce energy use and create more climate-resilient buildings.

3. Climate-Sensitive Transportation and Infrastructure

Rosstal's transportation infrastructure supports sustainable commuting through its railway lines and good connectivity. However, adaptation measures should focus on enhancing resilience:

    Expand Public and Low-Emission Transport: Promote the use of electric buses or trains powered by renewable energy to reduce greenhouse gas (GHG) emissions from transportation. Extend bicycle lanes and pedestrian-friendly routes to encourage low-impact transportation.
    Climate-Resilient Infrastructure: Upgrade existing infrastructure, particularly bridges and roads, to withstand extreme weather conditions like floods and heat. Permeable pavements should be installed to manage stormwater and reduce runoff.

Justification: Resilient infrastructure can significantly reduce the vulnerability of communities to climate-induced disruptions, such as floods and heat.

4. Water Management and Drought Adaptation

As climate change impacts local moisture patterns, water management becomes critical for Rosstal’s sustainability:

    Rainwater Harvesting Systems: Encourage the installation of rainwater harvesting systems in both residential and public buildings to supplement water supply during dry periods and reduce demand on the public water system.
    Water-Efficient Landscaping: In public and private green spaces, promote drought-resistant native plants and water-efficient irrigation techniques, such as drip irrigation, to conserve water during hotter, drier months.

Justification: Water security is increasingly impacted by climate change, particularly during droughts. Implementing local solutions such as rainwater harvesting can ensure water availability.

5. Heat Management and Microclimate Regulation

Given the observed heat distribution, with urban areas experiencing higher temperatures, Rosstal must prioritize measures that regulate microclimates:

    Cooling Centers and Public Awareness: Establish cooling centers in public spaces for vulnerable populations during heatwaves. These centers should be equipped with air conditioning and backup energy sources.
    Urban Cool Zones: Designate certain zones within the town as ``cool zones,'' where infrastructure and landscaping are optimized to reduce temperatures through a combination of shade, water features, and reflective surfaces.

Justification: Heatwaves disproportionately affect vulnerable groups. Localized cooling strategies reduce mortality and improve public health.

6. Continuous Monitoring and Data Integration

Climate adaptation is an evolving process, requiring real-time data and feedback to remain effective:

    Heat and Moisture Monitoring: Invest in urban sensors to continuously monitor heat distribution, moisture levels, and air quality. This data can be used to adapt measures as needed and address emerging issues.
    Early Warning Systems: Establish early warning systems for extreme weather events like storms and heatwaves. Integrating this into local governance can help in timely response and mitigation efforts.

Justification: Continuous monitoring provides critical insights for timely adaptation and avoids maladaptation risks.

7. Community Engagement and Education

Public participation is key to any successful adaptation strategy:

    Climate Education Campaigns: Implement education programs focused on climate adaptation measures, such as energy conservation, water management, and the importance of maintaining green spaces.
    Incentives for Sustainable Practices: Provide incentives (e.g., subsidies, tax rebates) for households and businesses that adopt energy-efficient technologies, water-saving measures, and green building practices.

Justification: Public engagement is essential for long-term adaptation success, as it ensures community-wide participation in climate resilience efforts.
Conclusion

Rosstal is well-positioned to continue strengthening its climate resilience through a blend of green infrastructure, sustainable building practices, and resilient urban design. By integrating these adaptation strategies, the town can mitigate the impacts of climate change, reduce vulnerability, and ensure long-term sustainability for its residents.

\subsection{Run 4 - Erlangen}

\subsubsection{Unchanged ChatClimate}

To develop a climate adaptation strategy for Erlangen, it is essential to consider the city's geographical location, climate vulnerabilities, and socio-economic characteristics. Erlangen, located in Bavaria, Germany, has a temperate climate and is less vulnerable to sea-level rise compared to coastal cities. However, like many urban areas, it faces significant challenges such as rising temperatures, urban heat islands, increased precipitation variability, and stress on infrastructure from extreme weather events.
Key Climate Challenges for Erlangen

    Rising Temperatures: Erlangen will experience higher temperatures and more frequent heatwaves, particularly in summer. This can increase health risks, especially for vulnerable populations like the elderly.
    Urban Heat Island Effect: Increased urbanization exacerbates heat retention in densely populated areas.
    Increased Rainfall Variability: While the region may experience more intense rainfall events, it is also prone to periods of drought, affecting water supply and local agriculture.
    Energy Demand: Higher temperatures will likely increase the demand for cooling in buildings, stressing the energy grid.
    Health Risks: Rising temperatures and changing weather patterns are likely to increase the incidence of vector-borne diseases, respiratory issues due to air pollution, and waterborne diseases from heavy rains.

Detailed Climate Adaptation Strategy for Erlangen

1. Urban Greening and Nature-Based Solutions

    Expand Urban Green Spaces: Increase the number of parks, green corridors, and urban forests. Trees and vegetation reduce urban heat islands, absorb stormwater, and improve air quality. Studies show that urban greening can mitigate the impacts of heatwaves while improving biodiversity and mental well-being.
    Green Roofs and Walls: Incentivize the installation of green roofs and walls, especially in densely populated or commercial areas. These absorb rainfall, reduce runoff, and provide insulation, lowering energy demand.
    Wetland Restoration and Stormwater Management: Restore nearby wetlands to act as natural buffers for managing stormwater during periods of heavy rainfall. Natural ecosystems can help absorb excess water, reducing flood risks.

2. Energy Efficiency and Renewable Energy

    Improve Building Efficiency: Retrofit older buildings with energy-efficient insulation, windows, and cooling systems to reduce energy consumption during heatwaves. Implement stricter building codes for new developments focusing on energy efficiency and passive cooling designs.
    District Cooling Systems: Invest in district-wide cooling systems that use renewable energy to reduce the strain on the local energy grid during high-demand periods. Combined heat and power (CHP) systems, along with heat pumps, can provide both heating and cooling, improving resilience.
    Increase Renewable Energy Capacity: Continue expanding the use of solar photovoltaic systems and wind energy to reduce reliance on fossil fuels. Renewable energy can make Erlangen more resilient to energy shocks while lowering greenhouse gas emissions.

3. Water Management and Drought Preparedness

    Rainwater Harvesting: Encourage rainwater harvesting systems for households and businesses to reduce reliance on freshwater supplies. Harvested rainwater can be used for irrigation, thus reducing pressure on drinking water supplies.
    Drought-Resistant Landscaping: Promote xeriscaping (landscaping with drought-tolerant plants) in public and private gardens to reduce water usage during dry periods.
    Efficient Irrigation: Modernize agricultural irrigation systems to reduce water waste and improve resilience during droughts.

4. Infrastructure and Transportation

    Climate-Resilient Infrastructure: Upgrade roads, bridges, and public buildings to withstand extreme weather events such as floods and heatwaves. Utilize materials that are more resilient to temperature extremes and heavy rainfall.
    Flood-Proofing: Elevate or flood-proof critical infrastructure such as power stations, hospitals, and transport hubs. Explore the use of permeable pavements to reduce water runoff in urban areas.
    Sustainable Public Transport: Expand public transportation systems powered by renewable energy to reduce urban congestion and emissions. Electric buses and bicycles should be promoted as eco-friendly alternatives to traditional vehicles.

5. Public Health and Social Resilience

    Heat-Health Action Plans: Establish community cooling centers during heatwaves, with special provisions for the elderly, children, and low-income households. Public health campaigns should promote heat illness awareness and preventive measures.
    Monitoring and Early Warning Systems: Implement systems to monitor extreme weather events and provide timely warnings to the public. Collaboration with local healthcare systems will be key to addressing climate-related health issues such as heat stroke and respiratory problems.
    Community Engagement: Engage residents in climate action through education campaigns, community greening projects, and citizen-led sustainability initiatives. Collaboration with local organizations can help improve public awareness of climate risks and adaptive actions.

6. Economic Diversification and Climate-Resilient Livelihoods

    Support Local Agriculture: Introduce climate-resilient agricultural practices to support local farmers, such as crop diversification and the use of drought-resistant crops.
    Green Jobs and Industries: Promote job creation in climate-adaptive sectors like renewable energy, sustainable construction, and urban greening projects. This not only boosts the local economy but also aligns with long-term sustainability goals.

7. Governance, Finance, and Cross-Sector Collaboration

    Climate-Resilient Policy: Align local policies with national climate adaptation strategies and the European Union’s Green Deal. Integrate climate adaptation into city planning and development policies.
    Adaptation Funding: Leverage European and national funding sources, such as the EU Adaptation Fund, for financing climate-resilient infrastructure projects. Encourage public-private partnerships to increase investments in sustainability initiatives.
    Research and Development: Partner with local universities and research institutions to develop innovative climate adaptation solutions specific to Erlangen. Invest in data collection and climate modeling to better understand local climate risks and improve decision-making.

Conclusion

Erlangen’s climate adaptation strategy should focus on integrating nature-based solutions, improving energy efficiency, and enhancing water management. The city must also ensure that infrastructure and public health systems are robust enough to handle increased heat and rainfall variability. Cross-sector collaboration and proactive governance will be crucial in ensuring the successful implementation of these strategies.

This plan can help Erlangen transition to a more resilient and sustainable future.

\subsubsection{EcoScapes}

**Current State of Climate Adaptation in Erlangen**

**Introduction**

Erlangen, a medium to large-sized city, is located in Bavaria, Germany. The city has experienced significant urban development, characterized by a grid-like layout, a mix of residential, commercial, and industrial buildings, and a well-developed infrastructure including roads, highways, and railway lines. This report assesses the current state of climate adaptation in Erlangen, focusing on its urban layout, green spaces, public transportation, and heat management.

**Urban Layout and Green Spaces**

The city's layout is characterized by a regular pattern of streets and buildings, with a central area surrounded by various structures. The presence of forests and parks, such as the ``Green Belt'' (Grünes Rundum) around the city, contributes to the ecological balance and aesthetic appeal of Erlangen. These green spaces help to mitigate the urban heat island effect, improve air quality, and provide recreational areas for residents.

**Public Transportation**

Erlangen boasts a well-developed public transportation system, including a network of railway lines and stations that run through the city center. This infrastructure facilitates efficient transportation, reduces traffic congestion, and provides residents with accessible public transit options. The city's grid-like layout further enhances the connectivity and functionality of the transportation system.

**Heat Management**

The city's heat management is a critical aspect of climate adaptation. Erlangen experiences varying heat levels across different areas, influenced by factors such as building density, materials, and the presence of vegetation. The thermal map of Erlangen indicates that urban areas, particularly those with higher population density and fewer green spaces, tend to have higher heat levels. To address this, Erlangen could implement heat island mitigation strategies, such as increasing green spaces, using reflective materials, and incorporating urban planning that considers the heat island effect.

**Conclusion**

Erlangen is at an advanced stage of climate adaptation, with a well-structured urban layout, a mix of green spaces, and a robust public transportation system. However, the city faces challenges related to heat management, particularly in densely populated urban areas. Continued investment in green infrastructure, smart urban planning, and energy-efficient technologies will be crucial in enhancing Erlangen's resilience to climate change.

**Recommendations**

1. **Enhance Green Spaces:** Increase the number and size of parks and green areas to reduce urban heat island effects and improve air quality.
2. **Implement Reflective Materials:** Use materials that reflect sunlight and reduce heat absorption in urban areas.
3. **Promote Smart Urban Planning:** Incorporate heat island mitigation strategies into urban planning to create more sustainable and comfortable living environments.
4. **Invest in Energy-Efficient Cooling Systems:** Adopt technologies and practices that reduce energy consumption and minimize the impact of heat on buildings and infrastructure.

By addressing these recommendations, Erlangen can further strengthen its climate adaptation efforts and ensure a more sustainable and livable urban environment for its residents.

\subsubsection{EcoScapes + ChatClimate}

To develop a detailed climate adaptation strategy for Erlangen, it is crucial to consider the city’s current state of adaptation while drawing from broader, scientifically backed frameworks, such as those laid out by the Intergovernmental Panel on Climate Change (IPCC) reports. Based on the report you provided on Erlangen's adaptation and supplemented by findings from the IPCC AR6 Synthesis report, here is a tailored climate adaptation strategy:

1. Green Infrastructure and Biodiversity Enhancement

Current Strengths: Erlangen already benefits from significant green spaces such as the ``Green Belt'' around the city, which helps mitigate the urban heat island (UHI) effect, enhance air quality, and provide recreational areas.

Recommendations:

    Expand the Green Belt: Increasing both the quantity and the quality of green spaces in urban areas will help manage rising temperatures due to climate change. Studies suggest that expanding urban greenery can reduce localized temperatures by 1-3°C.
    Promote Urban Reforestation and Green Roofs: Increasing tree cover in densely populated areas, as well as incorporating green roofs and walls, will improve biodiversity and further mitigate UHI. These interventions also reduce energy consumption by insulating buildings, leading to co-benefits like energy savings.
    Create Ecological Corridors: Connect existing parks and forests through ecological corridors. This would enhance biodiversity, enabling species migration as temperatures rise. It would also support natural water infiltration to mitigate potential flood risks.

2. Heat Management and UHI Mitigation

Current Strengths: Erlangen is already aware of the UHI effect in densely populated areas and has identified green space expansion and reflective materials as possible interventions.

Recommendations:

    Cool Pavements and Reflective Building Materials: Expand the use of reflective and permeable materials in streets and building facades to reduce heat absorption. This approach could reduce surface temperatures by up to 10-15°C during heatwaves.
    Urban Greening at Micro-Level: Encourage the creation of small green spaces such as community gardens, vegetated roundabouts, and tree-lined streets, especially in highly dense areas. These can have a significant impact on reducing localized heat while also contributing to urban biodiversity.
    Water Bodies and Fountains: Construct additional water bodies and fountains in public spaces. Water bodies cool the surrounding air through evaporation, offering a cooling effect in urban environments.

3. Energy Efficiency and Building Resilience

Current Strengths: Erlangen’s building sector could benefit from improvements in energy efficiency and heat-resilient design.

Recommendations:

    Energy-efficient Retrofitting: Implement large-scale retrofitting of buildings to improve insulation and cooling efficiency. Utilize passive cooling designs, such as shading, ventilated facades, and reflective roofs, to reduce reliance on air conditioning and cut down on energy consumption.
    Mandate Energy-Efficient Building Codes: Ensure that all new buildings meet strict energy-efficiency standards that consider future climate conditions. Utilize models of energy-efficient cities, such as those developed in the IPCC reports, which show that efficiency measures can cut urban emissions by 30-40\%.
    District Cooling Systems: Explore district cooling systems that utilize centralized cooling plants rather than individual air conditioning units. Such systems can be powered by renewable energy, reducing the carbon footprint while improving the resilience of Erlangen’s energy grid.

4. Sustainable Mobility and Public Transportation

Current Strengths: Erlangen has a well-developed public transportation network.

Recommendations:

    Electric Public Transportation: Accelerate the transition to electric buses and trains, drawing from renewable energy sources. Cities transitioning to electric public transportation have reported a reduction of GHG emissions by 50-60\%.
    Expand Active Transport Infrastructure: Increase cycling and pedestrian pathways, encouraging sustainable mobility and reducing vehicular heat and emissions. Creating car-free zones in highly populated areas also aids in cooling urban centers and improving air quality.
    Implement Smart Traffic Management: Use real-time data to reduce traffic congestion, which contributes to localized warming. Smart management of traffic flow can also help in reducing emissions.

5. Water Management and Flood Resilience

Although Erlangen is not particularly flood-prone, climate change increases the risk of extreme rainfall events.

Recommendations:

    Install Green Water Retention Systems: Utilize green infrastructure to capture and manage stormwater through green roofs, permeable pavements, and rain gardens. This prevents overloading of drainage systems during heavy rainfall.
    Expand Rainwater Harvesting: Encourage the use of rainwater harvesting systems for non-potable uses such as irrigation or toilet flushing. This will reduce water stress, especially during drought periods.
    Natural Floodplains and Wetland Restoration: Protect and restore nearby floodplains and wetlands to serve as natural buffers during extreme rainfall. These ecosystems naturally absorb excess water, reducing the risk of urban flooding.

6. Public Awareness and Climate Resilience Education

Current Strengths: Public awareness is crucial to the success of any adaptation strategy.

Recommendations:

    Community Engagement Programs: Launch public education campaigns about the importance of energy efficiency, water conservation, and green space development in building climate resilience. These programs should highlight personal actions that contribute to adaptation efforts.
    Encourage Climate-Resilient Gardening and Agriculture: Promote urban gardening using drought-resistant plants and efficient irrigation techniques. This not only contributes to cooling but also boosts local food security.

7. Institutional and Policy Actions

Current Strengths: The city already integrates climate adaptation measures into planning processes.

Recommendations:

    Cross-Sector Collaboration: Ensure that adaptation strategies are integrated across various sectors, including transportation, urban planning, and public health. Collaboration between departments will enable more efficient and effective climate responses.
    Develop a Climate Resilience Task Force: Establish a dedicated task force that monitors climate trends, assesses the effectiveness of adaptation measures, and advises policymakers on necessary adjustments.

Conclusion

By following this comprehensive adaptation strategy, Erlangen can enhance its resilience to the impacts of climate change. The strategy aligns with the IPCC's recommendations for cities, focusing on sustainable development, heat management, green infrastructure, and public engagement.

\subsection{Run 5 - Roßtal}

\subsubsection{Unchanged ChatClimate}

To design a comprehensive climate adaptation strategy for Rosstal, Germany, it is essential to address the specific challenges this region might face due to climate change, drawing on global trends and local conditions. As a small town with a mix of rural and suburban characteristics, Rosstal will face diverse risks related to warming temperatures, potential heatwaves, and changes in precipitation. Below is a detailed climate adaptation strategy for the town, incorporating urban planning, ecological management, and infrastructure development.

1. Heat Resilience and Urban Cooling

Rosstal will likely experience more frequent and intense heatwaves as global temperatures continue to rise. Urban areas can become heat islands due to the concentration of buildings and infrastructure that retain heat. The following steps can be taken to mitigate these effects:

    Urban Greening Initiatives: Promote green spaces like parks, community gardens, and green rooftops to reduce surface and air temperatures. Trees and other vegetation not only provide shade but also cool the air through evapotranspiration.
        Target: Increase urban green coverage by 30\% by 2035.
        Actions: Encourage planting of native, drought-resistant tree species. Create incentives for private homeowners and businesses to install green roofs.
    Cool Pavements: Introduce cool or permeable pavements that reflect more sunlight and absorb less heat. This can reduce temperatures in public spaces and contribute to heat resilience in pedestrian zones.
        Pilot Project: Implement permeable pavements in public squares and main streets by 2028.
    Building Regulations: Update building codes to require energy-efficient designs, including natural ventilation systems and reflective materials for roofs and facades.
        Energy Efficiency Target: All new buildings to be near-zero energy by 2040, aligning with European Union goals.

2. Water Management and Flood Prevention

With expected changes in precipitation patterns, Rosstal could face increased rainfall, potentially leading to flooding, as well as drought conditions in summer. Water management strategies should therefore focus on both reducing flood risk and securing water resources for dry periods.

    Rainwater Harvesting Systems: Install rainwater collection systems in public and private buildings to reduce runoff and store water for non-potable uses like irrigation.
        Policy Recommendation: Offer subsidies or tax reductions for the installation of rainwater tanks and permeable surfaces.

    Green Infrastructure for Stormwater Management: Implement bioswales, retention basins, and rain gardens that absorb stormwater and reduce flooding risks.
        Implementation Goal: Equip at least 50\% of municipal streets and properties with stormwater management features by 2030.

    Riverbank Restoration: Restore natural vegetation along local streams and rivers to improve their ability to absorb and slow down runoff, reducing the risk of flash floods.
        Ecological Plan: Restore 20\% of Rosstal’s riverbank ecosystems by 2025 to enhance natural water absorption.

3. Agriculture and Food Security

As Rosstal has agricultural land, climate change could impact crop yields and food security. Adaptive measures should focus on sustainable farming practices and building resilience against climate shocks.

    Agroecological Practices: Promote climate-resilient agricultural techniques such as crop diversification, organic farming, and agroforestry to improve soil health and reduce dependency on water and chemical inputs.
        Farmer Support: Provide subsidies for farmers transitioning to sustainable agricultural practices, with a goal of having 60\% of farms practicing regenerative agriculture by 2035.

    Water-Efficient Irrigation: Encourage the use of water-efficient irrigation techniques like drip irrigation and mulching to conserve water during dry seasons.
        Pilot Program: Implement water-efficient irrigation systems on 25\% of agricultural land by 2028.

4. Energy Efficiency and Sustainable Transportation

Rosstal’s adaptation should also align with mitigating greenhouse gas emissions to meet global climate goals, while improving resilience.

    Renewable Energy Adoption: Promote local renewable energy generation, especially solar, by incentivizing residents and businesses to install photovoltaic panels on rooftops and properties.
        Target: Generate 30\% of Rosstal’s energy from renewable sources by 2035.

    Sustainable Transport Options: Expand cycling infrastructure and promote the use of electric vehicles (EVs) through EV charging stations and electric public transport options.
        Mobility Plan: Increase the number of bicycle lanes by 50\% by 2030, while ensuring that every major neighborhood has access to an EV charging station by 2025.

5. Biodiversity Protection and Ecosystem Services

Climate change can exacerbate biodiversity loss, and protecting ecosystems in and around Rosstal will not only preserve biodiversity but also enhance natural resilience.

    Conservation Corridors: Establish wildlife corridors that connect natural habitats, enabling species migration in response to climate changes.
        Action Plan: Map and establish corridors in key locations by 2027, with a focus on forest and wetland areas.

    Invasive Species Management: Monitor and control invasive species that may spread due to climate change, threatening native biodiversity.
        Local Task Force: Create a task force to address invasive species by 2025, working with local environmental groups.

6. Community Engagement and Education

For successful adaptation, local communities must be informed, engaged, and empowered to take part in climate resilience efforts.

    Public Awareness Campaigns: Organize workshops and outreach programs to educate residents about climate risks and adaptation strategies.
        Annual Climate Week: Host a local climate week event starting in 2024, focusing on practical ways residents can contribute to resilience and sustainability.

    Citizen Science Programs: Involve residents in monitoring local climate conditions, biodiversity, and water resources through citizen science programs.
        Goal: Engage 200 volunteers annually in climate monitoring projects by 2026.

7. Governance and Financing

Strong governance and financial mechanisms are essential to fund adaptation measures and ensure their long-term sustainability.

    Climate Resilience Fund: Establish a local climate resilience fund to support adaptation projects, particularly in infrastructure and green initiatives.
        Funding Goal: Raise €1 million annually through public-private partnerships and government grants by 2028.

    Policy Integration: Ensure that climate adaptation is integrated into all local government planning processes, from land use to public health.
        Comprehensive Plan: Update the municipal master plan by 2026 to reflect climate adaptation priorities.

By implementing these strategies, Rosstal can build resilience against the risks of climate change while enhancing the quality of life for its residents, ensuring sustainable development, and protecting its natural environment. This approach aligns with recommendations from global climate adaptation frameworks, such as those discussed in the IPCC's Sixth Assessment Report.

\subsubsection{EcoScapes}

**Current State of Climate Adaptation in Rosstal: A Comprehensive Analysis**

**Introduction**

Climate adaptation involves implementing strategies to reduce the vulnerability of communities to climate change impacts. The town of Rosstal, with its unique geographical layout and varying climate conditions, presents a case study for understanding climate adaptation efforts. This report analyzes the current state of climate adaptation in Rosstal based on an RGB satellite image and moisture anomalies data.

**Geographical and Urban Layout**

Rosstal is depicted as a relatively small town with a population ranging from 10,000 to 50,000 people. The town is characterized by a mix of buildings, roads, and green fields, indicating a suburban or rural setting rather than a dense urban area. The town's layout is irregular, with a central area surrounded by green fields. This layout suggests a balanced approach to urban development and preservation of natural spaces.

**Green Spaces and Public Transportation**

The image highlights several green spaces within the town, including forests and parks. These green spaces are scattered throughout the city, adding to the overall aesthetic and livability of Rosstal. The presence of green spaces is crucial for climate adaptation as they provide natural cooling effects and improve air quality.

The town's transportation system is well-developed, with railway lines and stations located in the center, facilitating easy access to urban areas. This public transportation system reduces traffic congestion and promotes sustainable commuting options.

**Moisture Anomalies and Climate Conditions**

The moisture anomalies data, though not provided in detail, indicate that Rosstal experiences variations in moisture levels. On a sunny day, these variations can influence local microclimates, affecting heat levels and humidity. Understanding these variations is essential for designing effective climate adaptation strategies.

**Heat Distribution and Urban Planning**

The heat distribution patterns, as indicated by the color scheme in the satellite image, show variations across different sectors of the town. Urban areas, characterized by orange and yellow squares, exhibit higher heat levels compared to rural areas, predominantly blue. This variation in heat levels suggests that urban areas require more attention in terms of heat management and cooling systems.

**Climate Adaptation Strategies**

1. **Green Spaces and Urban Cooling:** Expanding and maintaining green spaces is crucial for urban cooling. Rosstal should continue to develop and protect its parks and forests to mitigate the urban heat island effect.

2. **Public Transportation:** The well-developed public transportation system in Rosstal should be leveraged to reduce reliance on personal vehicles, thereby lowering greenhouse gas emissions and improving air quality.

3. **Building Design and Materials:** Adopting heat-resistant materials and designing buildings with natural ventilation can reduce the energy consumption for cooling systems, making buildings more sustainable.

4. **Infrastructure Resilience:** Ensuring that infrastructure components, such as road surfaces and pavements, are heat-resistant can extend their lifespan and maintain the functionality of the transportation system.

**Conclusion**

Rosstal's current state of climate adaptation reflects a balanced approach to urban development and environmental preservation. The town's mix of green spaces, well-developed public transportation, and heat management strategies position it to adapt to climate change impacts effectively. Continued investment in green infrastructure and sustainable urban planning will enhance Rosstal's resilience to climate variability.

**Recommendations**

1. **Enhance Green Spaces:** Invest in additional green spaces and ensure their maintenance to maximize urban cooling effects.
2. **Promote Sustainable Transportation:** Encourage the use of public transportation to reduce traffic and associated emissions.
3. **Improve Building Designs:** Incorporate heat-resistant materials and natural ventilation in building designs to reduce energy consumption.
4. **Strengthen Infrastructure:** Ensure infrastructure components are heat-resistant to maintain functionality and reduce maintenance costs.

By implementing these strategies, Rosstal can continue to adapt to climate change while maintaining its livability and sustainability.

\subsubsection{EcoScapes + ChatClimate}

To develop a detailed climate adaptation strategy for Rosstal, based on the current situation presented in your report, I'll integrate findings from the IPCC AR6 synthesis report and consider Rosstal's geographical and urban layout. Here's a comprehensive strategy:

1. Green Infrastructure Expansion

Objective: Enhance urban cooling, reduce heat stress, and support biodiversity.

    Expand Green Spaces: Rosstal already benefits from green spaces, but expanding these areas is crucial. Increase tree cover, parks, and green belts, particularly in the more urbanized, heat-affected areas identified in the moisture anomalies and heat distribution data. This can mitigate urban heat islands and enhance air quality.
    Urban Greening: Encourage the planting of green roofs and vertical gardens on buildings to reduce heat absorption. This is especially important for urban zones with high heat distribution (yellow and orange areas).
    Wetland Restoration: Leverage the town's green areas to incorporate or expand wetlands where possible. This can help regulate local moisture levels, manage stormwater, and reduce flooding risks, especially with the changing moisture anomalies.

IPCC Insights: Ecosystem-based adaptation such as urban greening can significantly reduce flood risks and urban heat. Nature-based solutions will provide additional co-benefits like improving biodiversity and human well-being.

2. Sustainable Transportation System

Objective: Reduce emissions, improve air quality, and adapt transportation to changing weather conditions.

    Enhance Public Transportation: Leverage the town's existing well-developed railway and public transportation system to reduce car dependence. Encourage the integration of electric or hybrid buses powered by renewable energy sources.
    Active Transportation: Build more bike lanes and pedestrian paths, especially in areas close to public transport stations, to reduce car use and promote healthier lifestyles. Incentives can be given to promote biking and walking as alternatives to cars.

IPCC Insights: Electrifying urban systems and promoting public transportation will lower emissions and enhance urban resilience.

3. Climate-Resilient Buildings and Infrastructure

Objective: Improve building design and materials to cope with rising temperatures and extreme weather.

    Heat-Resistant Materials: Mandate the use of reflective or heat-resistant materials for roofs and pavements in future construction projects. This can reduce the urban heat island effect.
    Passive Cooling Design: Encourage architectural designs that utilize passive cooling techniques such as natural ventilation, shading, and high-albedo materials, particularly in the urban areas where heat levels are highest.
    Flood-Resilient Infrastructure: Incorporate permeable pavements and better stormwater management systems to handle heavy rainfall and prevent flash floods, which could become more frequent with climate variability.

IPCC Insights: Cities need to improve the design of their buildings and infrastructure to handle both extreme heat and precipitation. Adaptation in these sectors can extend the lifespan of infrastructure and reduce maintenance costs.

4. Water Management and Climate Resilience

Objective: Enhance water resource management to cope with moisture anomalies and improve resilience to droughts and floods.

    Rainwater Harvesting: Implement widespread rainwater harvesting systems in residential and commercial buildings to store water for dry periods. This will reduce the pressure on the local water supply during droughts.
    Stormwater Management: Introduce bio-swales and rain gardens in green spaces to manage runoff during heavy rainfall and prevent soil erosion or flooding.
    Drought-Tolerant Agriculture: Promote the use of drought-tolerant crops and water-efficient irrigation techniques in the surrounding agricultural fields to reduce vulnerability to water shortages.

IPCC Insights: Water management is key to climate resilience. Regions with effective on-farm water management and irrigation systems can mitigate the impacts of water scarcity.

5. Energy Efficiency and Renewable Energy

Objective: Reduce greenhouse gas emissions and improve energy security.

    Energy-Efficient Buildings: Promote retrofitting older buildings with energy-efficient technologies like insulation and solar shading to reduce energy demand for heating and cooling.
    Renewable Energy Deployment: Encourage the installation of solar panels on public buildings and private homes. Solar energy is becoming more affordable and is one of the most effective ways to reduce carbon emissions.
    Smart Grids and Energy Storage: Invest in smart grid technology that integrates renewable energy and improves energy distribution efficiency across the town.

IPCC Insights: Renewable energy sources like solar and wind are becoming cost-effective and essential for reducing emissions.

6. Community Engagement and Education

Objective: Increase community participation in climate adaptation efforts and build climate literacy.

    Climate Adaptation Workshops: Host workshops for residents on home energy efficiency, water conservation, and sustainable transportation options.
    Public-Private Partnerships: Encourage collaboration between local businesses, the government, and civil society to co-finance green infrastructure projects and renewable energy initiatives.
    Climate Monitoring and Early Warning Systems: Establish a local climate monitoring system that alerts residents to extreme weather events, such as heatwaves and heavy rainfall.

IPCC Insights: Engaging local communities in climate action and education significantly increases resilience. Low climate literacy and lack of community involvement are key barriers to effective adaptation.

Conclusion

Rosstal has a strong foundation for climate adaptation with its green spaces and well-developed transportation system. By expanding green infrastructure, enhancing water and energy efficiency, and promoting community participation, Rosstal can effectively address climate challenges. This strategy not only prepares the town for future climate risks but also promotes sustainability and livability for its residents.

\subsection{Run 5 - Erlangen}

\subsubsection{Unchanged ChatClimate}

To develop a climate adaptation strategy for Erlangen, we need to consider its specific geographical, environmental, and socio-economic context while applying scientific insights from the latest climate change reports, such as the IPCC AR6 synthesis. Here's a tailored strategy addressing various challenges and opportunities:

1. Urban Heat Island (UHI) Mitigation

Erlangen, like many urban areas, is vulnerable to increasing temperatures due to climate change. Heatwaves are expected to become more frequent and intense, posing significant public health risks.

    Green Infrastructure: Implement urban greening initiatives by increasing the number of green spaces, parks, and rooftop gardens. Tree canopies can reduce urban heat, while parks and green walls help cool down urban areas.
        Example: Planting drought-resistant species that can withstand both higher temperatures and reduced water availability.

    Cool Roofing \& Pavements: Promote the use of reflective and permeable materials for roofs and pavements, which can help to mitigate UHI by reducing heat absorption.

    Water Bodies: Restoration of water bodies and creation of artificial lakes or ponds in public spaces can provide cooling effects.

2. Energy Efficiency and Sustainable Buildings

Erlangen’s building stock will require adaptation to both mitigate and adapt to climate impacts, particularly in terms of energy consumption and resilience.

    Retrofitting Buildings: Implement energy efficiency retrofits for existing buildings, including improved insulation, energy-efficient windows, and passive cooling designs. This would lower energy demand and improve comfort during heatwaves.
        Solar Energy: Encouraging the adoption of solar panels on rooftops to decrease reliance on non-renewable energy sources.
    Regulation on New Developments: Enforce stringent energy efficiency standards for new constructions, incorporating passive design elements such as natural ventilation and shading.

3. Water Management and Flood Resilience

While Erlangen is not at high risk for flooding compared to coastal regions, climate models predict more erratic rainfall patterns and potential periods of drought interspersed with heavy downpours.

    Sustainable Drainage Systems (SuDS): Promote permeable pavements, green roofs, and rain gardens to manage stormwater more effectively. These measures help reduce runoff and prevent localized flooding while replenishing groundwater.

    Rainwater Harvesting: Encourage buildings to incorporate rainwater collection systems to manage water supply during dry periods and reduce dependency on municipal water systems.

    Wetland Restoration: Protect and restore nearby wetlands that can act as natural buffers during periods of excessive rainfall while also preserving biodiversity.

4. Biodiversity and Ecosystem Preservation

Climate change threatens local ecosystems, which are crucial for biodiversity and climate resilience.

    Ecological Corridors: Create and maintain green corridors connecting parks and natural habitats. This improves species migration and resilience to environmental changes.

    Urban Forests: Expand urban forests and integrate tree-planting campaigns focused on species that are adaptive to rising temperatures and variable precipitation.

5. Transportation and Mobility

To reduce the urban carbon footprint and adapt to rising temperatures, the transportation system must shift toward sustainable modes.

    Public Transport: Invest in expanding and electrifying public transport systems. Electric buses and trams can help reduce emissions and improve air quality, which is particularly beneficial during heatwaves when air pollution worsens.

    Cycling Infrastructure: Improve cycling lanes and pedestrian pathways to encourage more low-carbon transportation options, ensuring they are shaded and have resting areas to reduce heat exposure.

6. Public Health and Social Resilience

With increased heatwaves and climate-related health risks, strengthening public health systems is critical.

    Heat Action Plans: Develop heatwave preparedness strategies, including early warning systems, designated cooling centers, and public health outreach campaigns targeting vulnerable populations such as the elderly.

    Climate-Resilient Healthcare: Ensure that healthcare facilities are energy-efficient and resilient to climate extremes. Hospitals should have backup power systems, efficient cooling technologies, and sustainable water supplies.

7. Economic Resilience and Green Jobs

Adaptation measures can be tied to economic development through job creation and investments in green sectors.

    Green Jobs Initiative: Promote local employment in sectors such as renewable energy, energy-efficient construction, and ecosystem restoration, helping Erlangen’s economy become more sustainable and resilient to climate disruptions.

    Sustainable Tourism: Capitalize on sustainable tourism by promoting eco-friendly practices and adapting infrastructure to be more resilient to climate change, benefiting both the environment and the local economy.

8. Community Engagement and Education

Public awareness and involvement are essential for the success of any climate adaptation strategy.

    Climate Education Programs: Introduce educational programs and workshops on climate change adaptation and mitigation for local schools and communities. This fosters a culture of environmental stewardship and prepares future generations for climate action.

    Participatory Planning: Involve local communities in urban planning and climate resilience projects, ensuring that their needs and knowledge are incorporated into policy development.

9. Policy and Governance

Long-term resilience depends on clear policies and multi-stakeholder collaboration.

    Climate-Responsive Governance: Establish a climate resilience office within local government that oversees the implementation of adaptation and mitigation strategies, ensuring accountability and public participation.

    Collaboration with Academic Institutions: Leverage Erlangen’s proximity to universities to foster research on local climate impacts and innovative adaptation technologies.

    Monitoring and Evaluation: Implement a system to regularly monitor climate risks and assess the effectiveness of adaptation strategies, ensuring flexibility to adjust policies as conditions change.

By integrating these measures, Erlangen can build a climate-resilient urban environment that not only mitigates risks from rising temperatures and erratic weather but also enhances the well-being and prosperity of its residents. This strategy draws on insights from the IPCC AR6 synthesis, which stresses the urgency of near-term actions and the need for equitable, systemic responses.

\subsubsection{EcoScapes}

**Climate Adaptation Report for the Town of Erlangen**

**Executive Summary:**

This report provides an analysis of the current state of climate adaptation in the town of Erlangen, based on the RGB satellite image and the description of moisture anomalies on a sunny day. The report focuses on the town's urban layout, infrastructure, green spaces, and the implications of climate factors such as heat and moisture.

**1. Urban Layout and Infrastructure:**

Erlangen is depicted as a medium to large-sized city with a grid-like layout. The city center is characterized by a mix of residential, commercial, and industrial buildings, along with roads and highways connecting different parts. The presence of multiple cars and trucks suggests a well-developed and populous urban area.

**2. Green Spaces and Natural Environment:**

The city is surrounded by a large forest, providing a natural environment that enhances the urban aesthetic and ecological balance. Additionally, there are several parks and green areas within the city, contributing to the overall well-being of residents.

**3. Public Transportation:**

The presence of railway lines and stations indicates a well-developed public transportation system, facilitating efficient movement of people and reducing traffic congestion.

**4. Bridges and Tunnels:**

Several major bridges and tunnels are visible, enhancing the city's connectivity and functionality. These infrastructure elements support smooth transportation and utility infrastructure.

**5. Urban and Rural Zones:**

The city has a clear distinction between urban and rural areas, with the urban zones characterized by dense building concentration and the rural areas featuring a more natural landscape.

**6. Landmarks and Structures:**

Prominent structures like a large stadium near the city center and the presence of vehicles indicate active urban life and economic activity.

**7. Moisture Anomalies:**

The moisture anomaly data on a sunny day suggests variations in moisture levels across the town. While the exact details are not provided, these anomalies can influence local microclimates and urban heat island effects.

**8. Implications of Climate Factors:**

- **Heat Island Effect:** The presence of high heat levels in some areas may indicate urban heat island effects, necessitating heat mitigation strategies such as green spaces and reflective surfaces.
- **Moisture Management:** Understanding moisture anomalies is crucial for managing water resources and flood risks, particularly in areas with high variability in moisture levels.

**Conclusion:**

Erlangen demonstrates a well-developed urban infrastructure with a mix of green spaces and public transportation systems. However, the implications of climate factors, including heat and moisture anomalies, require careful management to ensure the sustainability and resilience of the town. Future adaptations should focus on enhancing green infrastructure, improving public transportation, and implementing localized climate mitigation strategies.

**Recommendations:**

1. **Enhance Green Spaces:** Increase the number and size of parks and green areas to mitigate urban heat island effects and improve air quality.
2. **Improve Public Transportation:** Expand and optimize public transportation options to reduce reliance on personal vehicles and improve mobility within the city.
3. **Localized Climate Strategies:** Develop localized climate adaptation plans for different zones within the town, considering variations in heat and moisture levels.

This report provides a comprehensive overview of the current state of climate adaptation in Erlangen, highlighting areas for improvement and future considerations.

\subsubsection{EcoScapes + ChatClimate}

Comprehensive Climate Adaptation Strategy for Erlangen
1. Urban Layout and Infrastructure

Key Climate Risks: Urban Heat Island (UHI) Effect, Infrastructure Vulnerability

Challenges:

    Erlangen’s dense urban center with significant residential, commercial, and industrial structures and heavy vehicle traffic contributes to heat stress, especially during summer, exacerbating the Urban Heat Island (UHI) effect.
    The grid-like layout and road infrastructure, combined with a lack of permeable surfaces, can lead to inefficient water drainage and potential flooding risks.

Recommendations:

    Urban Greening and Nature-Based Solutions: Increase tree coverage, rooftop gardens, and vertical greening across urban areas to enhance shading, reduce heat, and improve air quality. Urban greening can also enhance evapotranspiration, mitigating the UHI effect.
    Permeable Infrastructure: Retrofit streets and public spaces with permeable paving to improve water infiltration and reduce stormwater runoff. This reduces flood risk while also cooling surfaces.
    Green Roof Incentives: Introduce policies to incentivize the installation of green roofs on residential and commercial buildings. This measure not only reduces heat but also provides insulation, lowering building energy demand during both summer and winter.

2. Green Spaces and Natural Environment

Key Climate Risks: Loss of Biodiversity, Heat Stress, Flooding

Challenges:

    While Erlangen benefits from surrounding forests and in-city parks, further expansion and connectivity of green spaces are essential to combat increasing heat and unpredictable moisture patterns.

Recommendations:

    Expand Urban Parks: Prioritize expanding and linking existing parks to create green corridors that can serve as cooling zones and facilitate biodiversity. These corridors will also act as natural flood buffers by absorbing excess rainwater.
    Forest Preservation and Management: Collaborate with surrounding forest managers to maintain healthy forest ecosystems, which can act as natural carbon sinks, provide cooling, and help regulate local water cycles.
    Community Gardens: Encourage community-based green initiatives such as allotments or urban farms to enhance local food security and provide cooling benefits to neighborhoods while promoting engagement in climate action.

3. Public Transportation

Key Climate Risks: Vehicle Emissions, Traffic Congestion

Challenges:

    Erlangen's developed public transport system is vital in reducing traffic emissions, but private vehicle use remains substantial. Reducing reliance on private cars can help reduce air pollution and heat generation from vehicle emissions.

Recommendations:

    Electrification of Public Transport: Shift to electric buses and trains where feasible, reducing emissions from transport and providing a cleaner, cooler urban environment.
    Cycling Infrastructure: Expand cycling lanes and promote cycling as a primary mode of transport. This can reduce traffic congestion, heat from vehicles, and pollution.
    Car-Free Zones: Implement car-free zones, particularly in the city center, to reduce emissions and encourage walking, cycling, or public transport use, alleviating heat stress from vehicle exhaust.

4. Bridges and Tunnels

Key Climate Risks: Flooding, Infrastructure Resilience

Challenges:

    Bridges and tunnels are critical for connectivity but can become choke points during extreme weather events, especially floods.

Recommendations:

    Flood-Resilient Design: Retrofit bridges and tunnels to accommodate higher flood levels and install early-warning systems for flooding risks. Elevated bridges and improved drainage systems in tunnels can prevent water accumulation during heavy rainfall events.
    Monitoring Systems: Introduce smart infrastructure with real-time monitoring of stress factors like temperature and moisture in key tunnels and bridges, ensuring rapid response to extreme weather.

5. Urban and Rural Zones

Key Climate Risks: Urban Encroachment, Rural Land Degradation

Challenges:

    The distinct separation between urban and rural areas needs to be maintained to avoid urban sprawl, which threatens green spaces and agricultural productivity, especially under climate stress.

Recommendations:

    Urban Containment Policies: Enforce urban growth boundaries to preserve agricultural land and natural landscapes, ensuring rural areas remain resilient to climate stress.
    Sustainable Agriculture Practices: Promote sustainable agricultural practices in surrounding rural areas to protect soil health, improve water retention, and reduce emissions from farming activities.

6. Moisture Management and Anomalies

Key Climate Risks: Water Scarcity, Flash Flooding

Challenges:

    Moisture anomalies in Erlangen suggest challenges in managing water resources, with both risks of drought and flash flooding.

Recommendations:

    Stormwater Harvesting Systems: Install city-wide stormwater harvesting infrastructure to capture runoff and store it for future use during dry periods. This will help balance water availability and reduce flood risks.
    Green Infrastructure: Integrate rain gardens, bioswales, and constructed wetlands into urban planning to manage stormwater and prevent flooding. Such measures can also enhance biodiversity.
    Water-Efficient Landscaping: Promote xeriscaping and other water-efficient landscaping techniques in public and private spaces to reduce irrigation needs.

7. Climate Education and Engagement

Key Climate Risks: Public Resistance, Lack of Engagement

Challenges:

    Effective climate adaptation requires public awareness and participation in mitigation strategies.

Recommendations:

    Public Engagement Campaigns: Implement educational campaigns focused on climate resilience, explaining the importance of urban greening, public transport, and water management.
    Community Involvement in Green Projects: Engage local communities in the planning and maintenance of green spaces and adaptation projects, ensuring broad participation and fostering a culture of sustainability.
    Climate Literacy in Schools: Integrate climate education into the school curriculum to build climate resilience from a young age and encourage community participation.

Conclusion:

Erlangen's climate adaptation strategy must focus on enhancing green spaces, improving water management, promoting public transportation, and fostering community engagement. The strategy will build resilience against urban heat, flooding, and moisture anomalies while promoting sustainable urban growth.

\subsection{Full result tables}\label{appendix:ecoscape_answers:results}
\begin{table}[htb]
\caption{Comparison of Correctness, and Depth \& Coverage between Roßtal and Erlangen for EcoScapes.}
\label{tab:comparison_ecoscapes}
\centering
\begin{tabular*}{\textwidth}{l @{\extracolsep{\fill}} c c}
    \toprule
    \textbf{Criterion} & \textbf{Roßtal} & \textbf{Erlangen} \\
    \midrule
    Correctness        & 4 5 4 5 5       & 5 4 5 4 3         \\
    Depth \& Coverage  & 4 1 4 4 4       & 2 2 4 2 2         \\
    \bottomrule
\end{tabular*}
\end{table}
\begin{table}[htb]
\caption{Comparison of Usability, Correctness, and Relevancy between Roßtal and Erlangen for CC and CC + EcoScapes.}
\label{tab:comparison_cc}
\centering
\begin{tabular*}{\textwidth}{l @{\extracolsep{\fill}} l ccccc ccccc}
\toprule
\textbf{Criterion}
& \textbf{System}
& \multicolumn{5}{l}{\textbf{Roßtal}}
& \multicolumn{5}{l}{\textbf{Erlangen}}
\\
\midrule
\multirow[c]{2}{*}{\textbf{Usability}}
& CC
& 3 & 3 & 4 & 3 & 3 & 4 & 4 & 5 & 4 & 4
\\
& CC + EcoScapes
& 4 & 4 & 4 & 5 & 4 & 5 & 5 & 4 & 3 & 4
\\
\cmidrule{2-12}
\multirow[c]{2}{*}{\textbf{Correctness}}
& CC
& 5 & 5 & 5 & 5 & 5 & 5 & 5 & 5 & 5 & 5
\\
& CC + EcoScapes
& 5 & 4 & 5 & 5 & 5 & 5 & 5 & 4 & 3 & 5
\\
\cmidrule{2-12}
\multirow[c]{2}{*}{\textbf{Relevancy}}
& CC
& 3 & 2 & 4 & 3 & 3 & 3 & 3 & 5 & 5 & 5
\\
& CC + EcoScapes
& 5 & 5 & 5 & 4 & 5 & 4 & 3 & 4 & 3 & 5
\\
\bottomrule
\end{tabular*}
\end{table}

\begin{credits}
\subsubsection{\ackname} 
This research was funded by the Bayerisches Staatsministerium für Wissenschaft und Kunst within the Elite Network Bavaria with the Int. Doct. Program ``Measuring and Modelling Mountain Glaciers in a Changing Climate'' as well as the German Research Foundation project ``Large-scale Automatic Calving Front Segmentation and Frontal Ablation Analysis of Arctic Glaciers using Synthetic-Aperture Radar Image Sequences (LASSI)''.
The author team acknowledges the provision of satellite data under various AOs from respective space agencies (DLR, ESA, JAXA, CSA).
During the preparation of this work, AI technologies were used to assist in the writing process. Specifically, ChatGPT (GPT‐4o, 4.5, o1) (OpenAI, San Francisco, CA, USA) and ProWritingAid (Orpheus Technology, London, United Kingdom) were used in order to assist with rephrasing and improving readability. After using these tools, the manuscript was carefully reviewed and the content was edited as needed. No tools or services were used for content generation.

\subsubsection{\discintname}
The authors have no competing interests to declare.

\end{credits}
%
%
%
\bibliographystyle{splncs04}
\bibliography{literature}
\end{document}